%% file: acl_latex.tex
\documentclass[11pt]{article}

\usepackage[preprint]{acl}

\usepackage{times}
\usepackage{latexsym}
\usepackage{xurl}
\usepackage{philex}
\usepackage{siunitx}
\usepackage{array}
\usepackage[T1]{fontenc}

\usepackage[utf8]{inputenc}

\usepackage{microtype}

\usepackage{inconsolata}

\usepackage{amsmath}
\usepackage{amsfonts}
\usepackage{graphicx}
\usepackage{cleveref} 
\usepackage{booktabs, multirow} 
\usepackage{tabularx}
\usepackage{soul}
\usepackage{xcolor,colortbl} 
\usepackage{changepage,threeparttable} 
\usepackage{subcaption}
\usepackage{hyperref}
\usepackage{adjustbox}

\title{Rethinking the Idiomaticity Decomposability Hypothesis: Evidence from Distributional Learning}

\newcommand{\uos}{1}
\newcommand{\bht}{2}
\newcommand{\exeter}{3}
\newcommand{\rio}{4}

\author{Maggie Mi$^{\uos}~\;~$Golzar Atefi$^{\bht}$~\;~Atsuki Yamaguchi$^{\uos}$~\;~Felix Alexander Gers$^{\bht}$\\
\textbf{Aline Villavicencio$^{\uos,\exeter,\rio}$~\;~Nafise Sadat Moosavi$^{\uos}$}\\
$^{\uos}$University of Sheffield~\;~$^{\bht}$Berliner Hochschule für Technik (BHT)\\
$^{\exeter}$University of Exeter~\;~$^{\rio}$Federal University of Rio Grande do Norte, Brazil\\
\texttt{\{zmi1, ayamaguchi1, a.villavicencio, n.s.moosavi\}@sheffield.ac.uk}\\
\texttt{\{golzar.atefi, FelixAlexander.Gers\}@bht-berlin.de}}

\begin{document}
\maketitle
\begin{abstract}
Idioms can be analysed in terms of their decomposability, the extent to which constituent meanings contribute to the figurative whole. Decomposability is thought to predict syntactic flexibility. Usage-based accounts instead attribute idiom behaviour to distributional experience, such as speaker familiarity and predictability. We examine these views using contextualised language models as controlled distributional learners. We propose a model-internal measure of decomposability and relate it to human ratings, syntactic flexibility, and predictability while tracking idiom learning during pretraining. Model-derived decomposability correlates weakly with human judgments and shows a small but consistent negative relationship with syntactic flexibility. Pretraining analyses show that stabilisation of idiom representations in models is not explained by frequency alone. Instead, surprisal, decomposability, and frequency all contribute, with decomposability showing the strongest training-dependent effect.

\vspace{.2em}
\noindent\hspace{1.25em}%
\includegraphics[width=1.25em,height=1.25em]{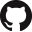}%
\hspace{.75em}%
\parbox[t]{\dimexpr\linewidth-1.25em-1.25em-.75em\relax}{%
  \raggedright
  \url{https://github.com/mi-m1/idiom_decomp}%
}

\end{abstract}

\section{Introduction}
\input{sections/1-introductions}

\section{Related Work}
\input{sections/2-related_work}

\section{Formalisations}
\input{sections/3-formulisations}

\section{Experimental Set-up}
\input{sections/4_experiments}

\section{Decomposability Across Humans, Models, and Syntax}\label{sec:human-model-alignment}
\input{sections/5_results}

\section{Acquisition Over-time}\label{sec:aot}
\input{sections/aot}

\section{Conclusion}\label{sec:dicscussion}
\input{sections/6-discussion}

\section*{Limitations}\label{sec:limitations}
\input{sections/8-limitations}

\section*{Acknowledgments}
We thank the reviewers for their comments and feedback. We acknowledge IT
Services at The University of Sheffield and the University of Oxford Advanced Research Computing (ARC) for the
provision of services for High Performance Computing. Finally, we thank Agata Savary and members of the MWE community for their discussions.

MM is supported by the UKRI AI Centre for Doctoral Training in Speech and Language Technologies (SLT) and their Applications funded by UK Research and Innovation [grant number EP/S023062/1]. For the purpose of open access, we have applied a Creative Commons Attribution (CC BY) licence to any Author Accepted Manuscript version arising. GA is supported by the Deutsche Forschungsgemeinschaft (DFG, German Research Foundation) – FIP-12 – [Project-ID 528483508], and the European Union [grant number 101079894]. AY is supported by the Engineering and Physical Sciences Research Council (EPSRC) [grant number EP/W524360/1] and the Japan Student Services Organization (JASSO) Student Exchange Support Program (Graduate Scholarship for Degree Seeking Students). AV's research is partly supported by UKRI (grants MR/U506734/1 and EP/T02450X/1), CNPq (406926/2025-5) and EQUATE.



\bibliography{custom}

\appendix

\section{Decomposability Score Derivation}
\label{sec:appendix}

\subsection{Examples of Derivation}
\label{sec:decomp_token_example}

In the following subsections, we demonstrate a worked example showing token-level $\Delta$ values for two canonical cases (e.g., rule of thumb (non/less decomposable idiom) vs. food for thought (more decomposable idiom). 

It is important to note that, throughout our analysis, we use the ranked position of each idiom as an indicator of its degree of decomposability, rather than relying on the raw values themselves.

\subsubsection{``rule of thumb''}
\textbf{Sentence:} 
The rule of thumb for working out maximum heart rate (MHR) is $225$ minus your age in years.

\textbf{Sentence glossed:}
The general principle for working out maximum heart rate (MHR) is $225$ minus your age in years.

\textbf{Masked Variants:}
\begin{itemize}
    \item \texttt{[CLS] the [MASK] of thumb for working out maximum heart rate (MHR) is 225 minus your age in years. [SEP]}
    \item \texttt{[CLS] the rule [MASK] thumb for working out maximum heart rate (MHR) is 225 minus your age in years. [SEP]}
    \item \texttt{[CLS] the rule of [MASK] for working out maximum heart rate (MHR) is 225 minus your age in years. [SEP]}
\end{itemize}

\vspace{0.4em}
\textbf{Token-level Importance ($\Delta_j$) - sorted.}
\begin{align*}
\text{rule}_{\Delta}  &= 0.0774 \\
\text{thumb}_{\Delta} &= 0.0666 \\
\text{of}_{\Delta}    &= 0.0663
\end{align*}

\textbf{Aggregation.} sum

\vspace{0.4em}
\textbf{Idiomatic Decomposability Score.} $0.2103$ 
\emph{(more transparent to figurative meaning)}

\subsubsection{``food for thought''}

\textbf{Sentence.} 
It won't do any harm, but I'd rather not give him food for thought, because I consider him an idiot and I don't think he's capable of interpreting it correctly.

\textbf{Sentence glossed.}
It won't do any harm, but I'd rather not give him anything that should be thought about, because I consider him an idiot and I don't think he's capable of interpreting it correctly.

\vspace{0.4em}
\textbf{Masked Variants.}
\begin{itemize}
    \item \texttt{... rather not give him [MASK] for thought, because I consider him an idiot ...}
    \item \texttt{... rather not give him food [MASK] thought, because I consider him an idiot ...}
    \item \texttt{... rather not give him food for [MASK], because I consider him an idiot ...}
\end{itemize}

\vspace{0.4em}
\textbf{Token-level Importance ($\Delta_j$) - sorted.}

\begin{align*}
\text{thought}_{\Delta} &= 0.0137 \\
\text{for}_{\Delta}     &= 0.0105 \\
\text{food}_{\Delta}    &= 0.0082
\end{align*}

\textbf{Aggregation.} sum

\vspace{0.4em}
\textbf{Idiomatic Decomposability Score.} $0.0324$ 
\emph{(less transparent to figurative meaning)}

\subsection{Aggregation}
\label{sec:agg_func}
The formulations of our aggregation metrics, for calculating expression-level decomposability score.

Mean
\begin{equation}
Decomp_{\mathrm{mean}} = \frac{1}{n} \sum_{j=1}^{n} \Delta_j .
\end{equation}

Maximum
\begin{equation}
Decomp_{\mathrm{max}} = \max_{1 \le j \le n} \Delta_j .
\end{equation}

To define dispersion-based measures, we first normalise the contributions:
\begin{equation}
p_j = \frac{\Delta_j}{\sum_{i=1}^{n} \Delta_i}, \quad \sum_{j=1}^{n} p_j = 1.
\end{equation}

Gini Dispersion
\begin{equation}
Decomp_{\mathrm{Gini}} = 1 - \sum_{j=1}^{n} p_j^2 .
\end{equation}

Entropy
\begin{equation}
Decomp_{\mathrm{Ent}} = - \sum_{j=1}^{n} p_j \log p_j .
\end{equation}

\section{Syntactic flexibility computation}
\label{syntactic flexibility example}
We provide an example for our computation of syntactic flexibility for the idiom \textit{break somebody's heart}. We first extract the frequencies of the idiom in each constructional type using the corresponding queries shown in Table \ref{tab:frequencies}.

\newcolumntype{P}{>{\footnotesize}p{0.20\textwidth}}
\begin{table}[!ht]
    \centering
    \footnotesize
    \setlength{\tabcolsep}{4pt} 
    \resizebox{\columnwidth}{!}{%
    \begin{tabular}{l P l l}
        \hline
        Construction & Query (lemmas) & Frequency & Probability\\
        \hline
         Base form & break [N/P]'s heart & 84758 & 0.75 \\
         Adjective Insertion & break [N/P]'s [ADJ] heart & 1719 & 0.01 \\
         Adverb Insertion & [ADV] break [N/P]'s heart & 13092 & 0.12\\
         Nominalization & breaking of [N/P]'s heart & 15 & 0.00\\
         Passive & [N/P]'s heart be break & 13526 & 0.12\\
         \hline
    \end{tabular}
    }
    \caption{Frequencies of the five constructional patterns of the idiom \textit{break somebody's heart} and their corresponding search queries. Except for the action nominalization \textit{breaking}, the verb was lemmatized in the searches to cover inflection and tense variation, and the slot somebody was implemented as a noun/pronoun search.}
    \label{tab:frequencies}
\end{table}

We next obtain the probabilities by dividing each individual frequency by the total frequency. The syntactic flexibility is then measured using Shannon entropy:

\begin{align*}
H(\text{\tiny break somebody's heart}) = - (
& 0.75 \log 0.75 \\
&+ 0.01 \log 0.01 \\
&+ 0.12 \log 0.12 \\
&+ 0.12 \log 0.12
) \approx 0.77
\end{align*}


\section{Experimental Details}

\subsection{Models}
\label{sec:model_info}

We use a total of 8 language models in this work. \Cref{tab:model-comparison} provides information regarding models used this paper.
\begin{table}[!ht]
\centering
\resizebox{0.45\textwidth}{!}{%
\begin{tabular}{lccc}
\hline
Model & Size & Tokens & Layers \\
\hline
bert-base-cased      & 110M & 30K  & 12 \\
bert-base-uncased    & 110M & 30K  & 12 \\
bert-large-cased     & 336M & 30K  & 24 \\
bert-large-uncased   & 336M & 30K  & 24 \\
ModernBert-base      & 149M & 2T   & 22 \\
ModernBert-large     & 395M & 2T   & 28 \\\midrule
Olmo-2-1124-7B       & 7B   & 4T   & 32 \\
Olmo-3-1025-7B       & 7B   & 5.93T& 32 \\
\hline
\end{tabular}
}
\caption{Comparison of language models used in this study.}
\label{tab:model-comparison}
\end{table}

\subsection{Datasets}
\label{sec:app_Datasets}

We provide an breakdown of the datasets used this in work in \Cref{tab:dataset-summary}. The idioms in IMPLI fall primarily into the functional categories listed in \Cref{tab:idiom_types_breakdown}.

\begin{table}[!htp]
\centering
\footnotesize
\begin{tabular}{l r | @{\hspace{1.5em}} l r}
\toprule
Category & \textit{n} & Category & \textit{n} \\
\midrule
VP   & 284 & ADJP        & 15 \\
PP   & 127 & ADVP        & 10 \\
NP   &  85 & OTHER (NUM) &  3 \\
     &      & S           &  2 \\
     &      & OTHER (INTJ)&  1 \\
\bottomrule
\end{tabular}
\caption{Distribution of idiomatic expressions across coarse-grained syntactic categories in IMPLI. We use the Bulkes and Tanner norms dataset \citep{Bulkes2017} for its human decomposability ratings. Specifically, we focus on the subset of idioms that overlap with IMPLI to enable direct comparison. The summary of the datasets are gathered in \Cref{tab:dataset-summary}.
}
\label{tab:idiom_types_breakdown}
\end{table}

\begin{table*}[ht]
\centering
\resizebox{\linewidth}{!}{%
\small
\begin{tabular}{lccp{0.45\textwidth}}
\hline
Dataset & \# Samples & \# Unique idioms & Example \\
\hline
IMPLI & 527 & 382 &
\small
Sentence ($S$): How have you weathered the storm? \newline
Paraphrase ($S_g$): How have you succeeded in getting through the difficult situation? \\\hline
Bulkes \& Tanner (subset) & 90 & 90 &
\small
Be a dark horse $\rightarrow$ 0.09 
\newline
Go back to basics $\rightarrow$0.97 \\
\hline
\end{tabular}%
}
\caption{Summary of datasets used in this study. \textsc{IMPLI} consists of idiomatic sentences paired with paraphrase. Bulkes \& Tanner dataset provides idioms annotated with human ratings across five dimensions (Familiarity, Predictability, Global Decomposability, Meaningfulness, and Literal Plausibility).}
\label{tab:dataset-summary}
\end{table*}

\subsection{Model-Human Proxies}
\label{sec:model-human-proxies}
We present an illustration of the model and human equivalent features in \Cref{tab:measures-human-vs-model}. 

\begin{table*}[t]
\centering
\small
\setlength{\tabcolsep}{6pt}
\renewcommand{\arraystretch}{1.15}
\begin{tabularx}{\textwidth}{@{}lXX|X@{}}
\toprule
\textbf{Human measure} & \textbf{Instruction} & \textbf{Response metric} & \textbf{Model Equivalent} \\
\midrule
Familiarity &
Rating how often Participants hear or use an idiom. &
Likert rating. & Frequency (\cref{subsec:frequency})\\
Predictability &
Sentence-completion (cloze): providing the first word that comes to mind. &
Proportion of responses matching the expected item. & Predictability (\cref{subsec:predictability})\\
Global decomposability &
Classification of idioms whose component parts contribute to their overall meaning. &
Binary categorization. & Decomposability (\cref{subsec:decomp})\\
\bottomrule
\end{tabularx}
\caption{Human-elicited measures (adapted from \citet{Bulkes2017}) and their model-derived counterparts.}
\label{tab:measures-human-vs-model}
\end{table*}

\section{Robustness Analyses}
\label{sec:robust}
\subsection{Bootstrap Confidence Intervals for Model–Human Correlations}
\label{sec:robust_bootstrap}
To assess the stability of the observed correlation, we conducted a bootstrap resampling analysis on the best-performing configuration (BERT-large, final-layer representations, Wasserstein distance, sum-based aggregation). Given the modest sample size (n = 90), this procedure provides a more robust estimate of variability. The resulting 95\% confidence interval for the correlation coefficient was [0.07, 0.40]. As this interval excludes zero, the effect is unlikely to be attributable to sampling noise, although the width of the interval indicates substantial uncertainty in the magnitude of the relationship.

\subsection{Partial Correlation between Predictors}
\label{sec:robust_partial_corr}


\Cref{tab:partial_corr} presents the partial correlation matrices for the model and human datasets. Most partial correlations are small in magnitude, indicating that the predictors are largely distinct. However, a few significant correlations, especially those involving frequency, suggest some modest small relationships among the variables.

\begin{table*}[h!]
\centering
\begin{tabular}{lcccc}
\toprule
 & \multicolumn{2}{c}{Model Data} & \multicolumn{2}{c}{Human Data} \\
\cmidrule(lr){2-3} \cmidrule(lr){4-5}
 & Predictability & Frequency  
 & Predictability & Frequency \\
\midrule
Predictability  & 1.000 & - & 1.000 & - \\
Frequency       & 0.262* & 1.000 & -0.148 & 1.000\\
Decomposability & 0.050 & -0.236* & -0.085 & -0.356* \\ 
\bottomrule
\end{tabular}
\caption{Partial correlation matrices for human and model data. * denotes significance.}
\label{tab:partial_corr}
\end{table*}

\subsection{Pearson Correlation between Acquisition Predictors}
\label{sec:robust_pearson_corr_acquisition}

The Pearson correlations among log frequency, surprisal, and decomposability are small in magnitude, suggesting minimal multicollinearity among the regression predictors. These relationships are presented in \Cref{tab:pearson_corr}.

\begin{table}[h!]
\centering
\resizebox{\columnwidth}{!}{
\begin{tabular}{lccc}
\toprule
 & Log\_Frequency & Surprisal & Decomp \\
\midrule
Log\_Frequency & 1.000 & -0.066 & -0.005 \\
Surprisal      & -0.066 & 1.000 & -0.188 \\
Decomp         & -0.005 & -0.188 & 1.000 \\
\bottomrule
\end{tabular}
}
\caption{Pearson correlation matrix among regression predictors.}
\label{tab:pearson_corr}
\end{table}

\subsection{Variance Inflation Factors}
\label{sec:vif}

The variance inflation factors (VIFs) were examined to assess potential multicollinearity among the predictors included in the regression models. As shown in \Cref{tab:vif}, all VIF values are close to 1 for both the model data and the human data, indicating that multicollinearity is negligible. This suggests that the predictors contribute relatively independent information and that the regression estimates are unlikely to be distorted by strong linear relationships among the explanatory variables.

\begin{table}[h!]
\centering
\begin{tabular}{lrr}
\toprule
Predictor & Model Data & Human Data\\
\midrule
Predictability & 1.074 & 1.023\\
Log\_Frequency      & 1.0134 & 1.164\\
Decomposability & 1.059 & 1.146\\
\bottomrule
\end{tabular}
\caption{Variance Inflation Factors (VIFs).}
\label{tab:vif}
\end{table}

\section{Additional Results}
\subsection{Layer-wise Correlation}
\label{sec:layer-wise-corr-syntax}

\begin{figure*}[t]
    \centering
    \includegraphics[width=0.9\textwidth]{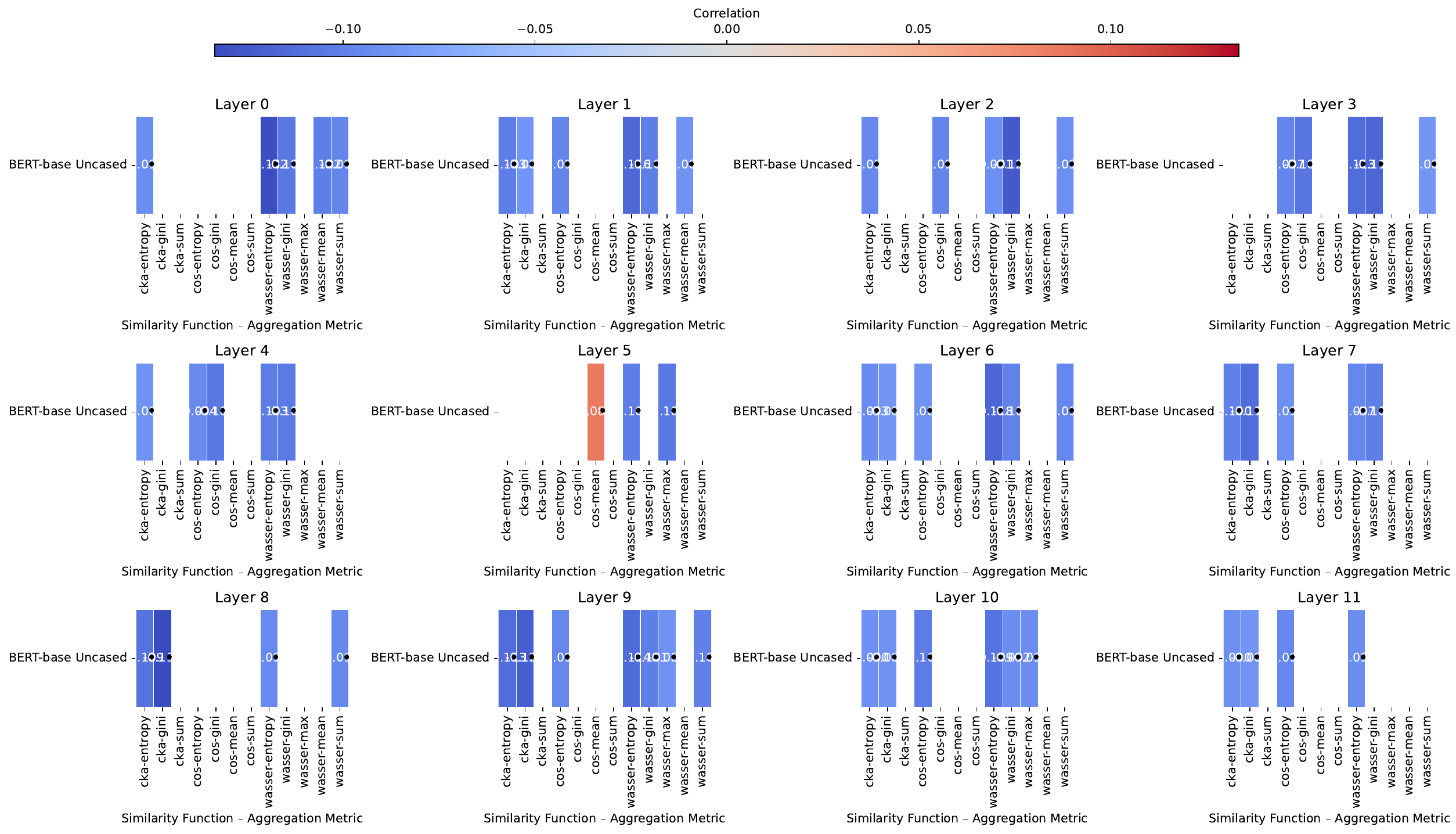}
    \caption{Correlation results for BERT-base Uncased}
    \vspace{-5pt}
    \label{fig:correlation_bert_base_uncased}
\end{figure*}

\begin{figure*}[t]
    \centering
    \includegraphics[width=0.9\textwidth]{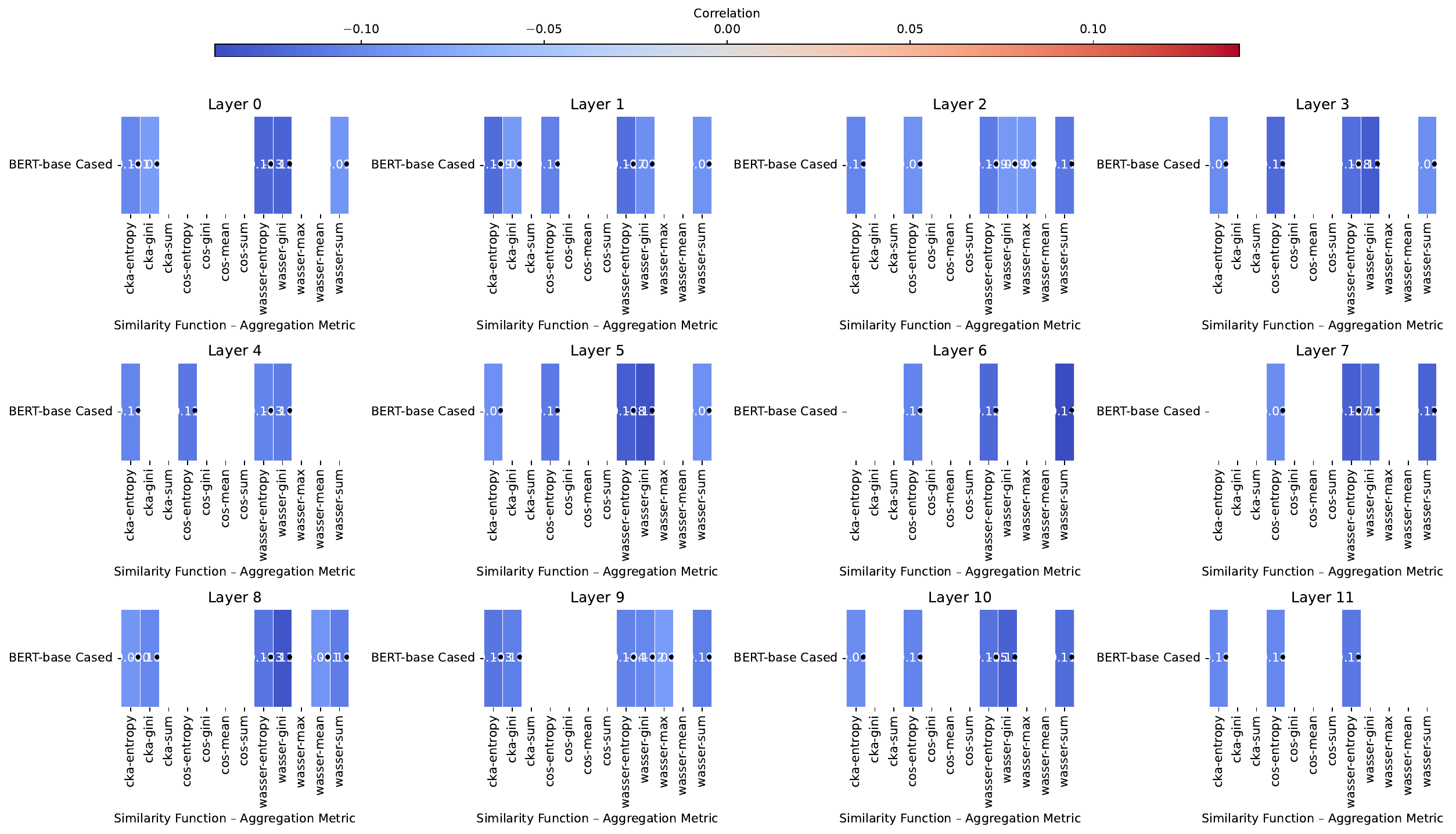}
    \caption{Correlation results for BERT-base Cased}
    \vspace{-5pt}
    \label{fig:correlation_bert_base_cased}
\end{figure*}

\begin{figure*}[t]
    \centering
    \includegraphics[width=0.9\textwidth]{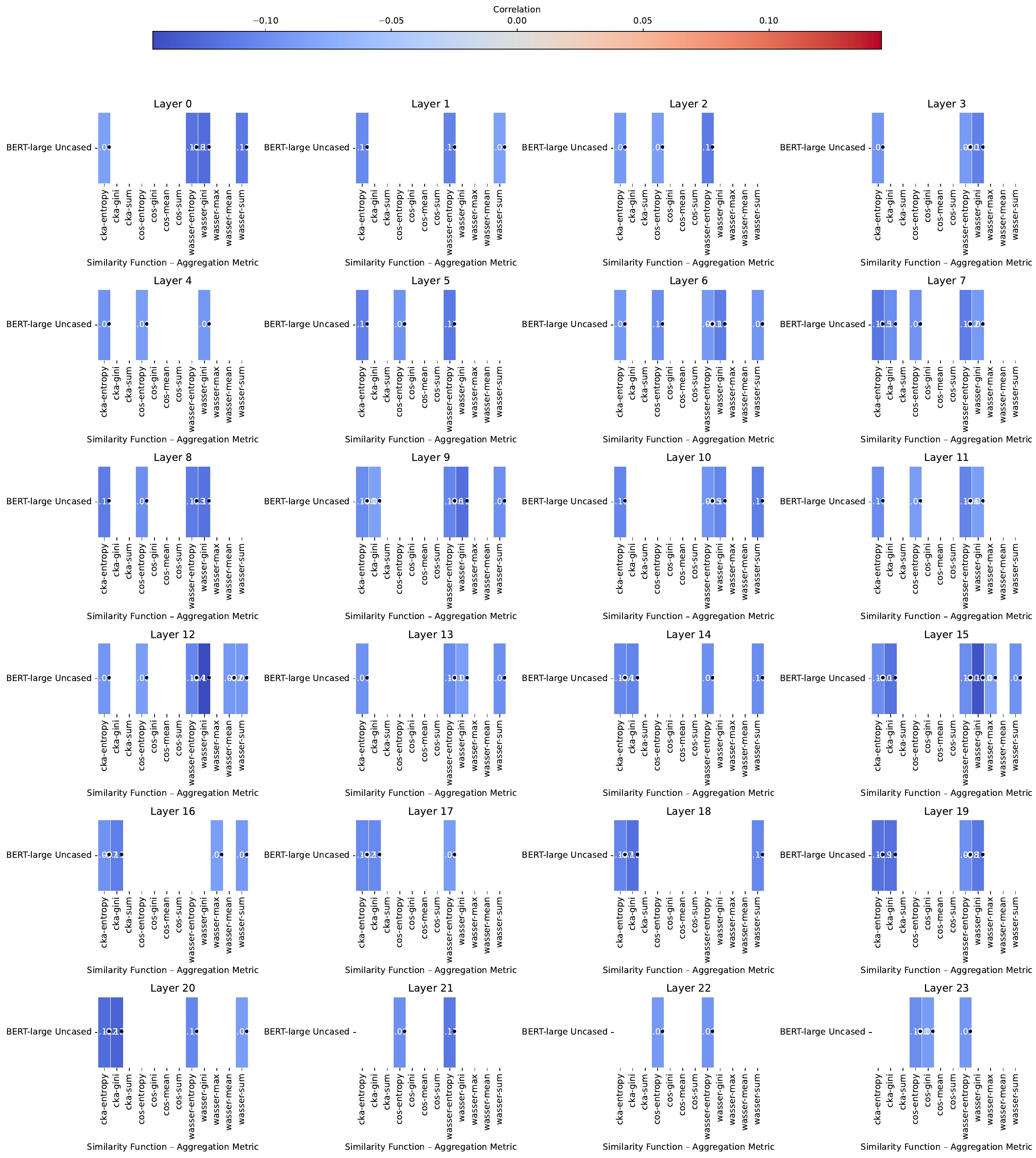}
    \caption{Correlation results for BERT-large Uncased}
    \vspace{-5pt}
    \label{fig:correlation_bert_large_uncased}
\end{figure*}

\begin{figure*}[t]
    \centering
    \includegraphics[width=0.9\textwidth]{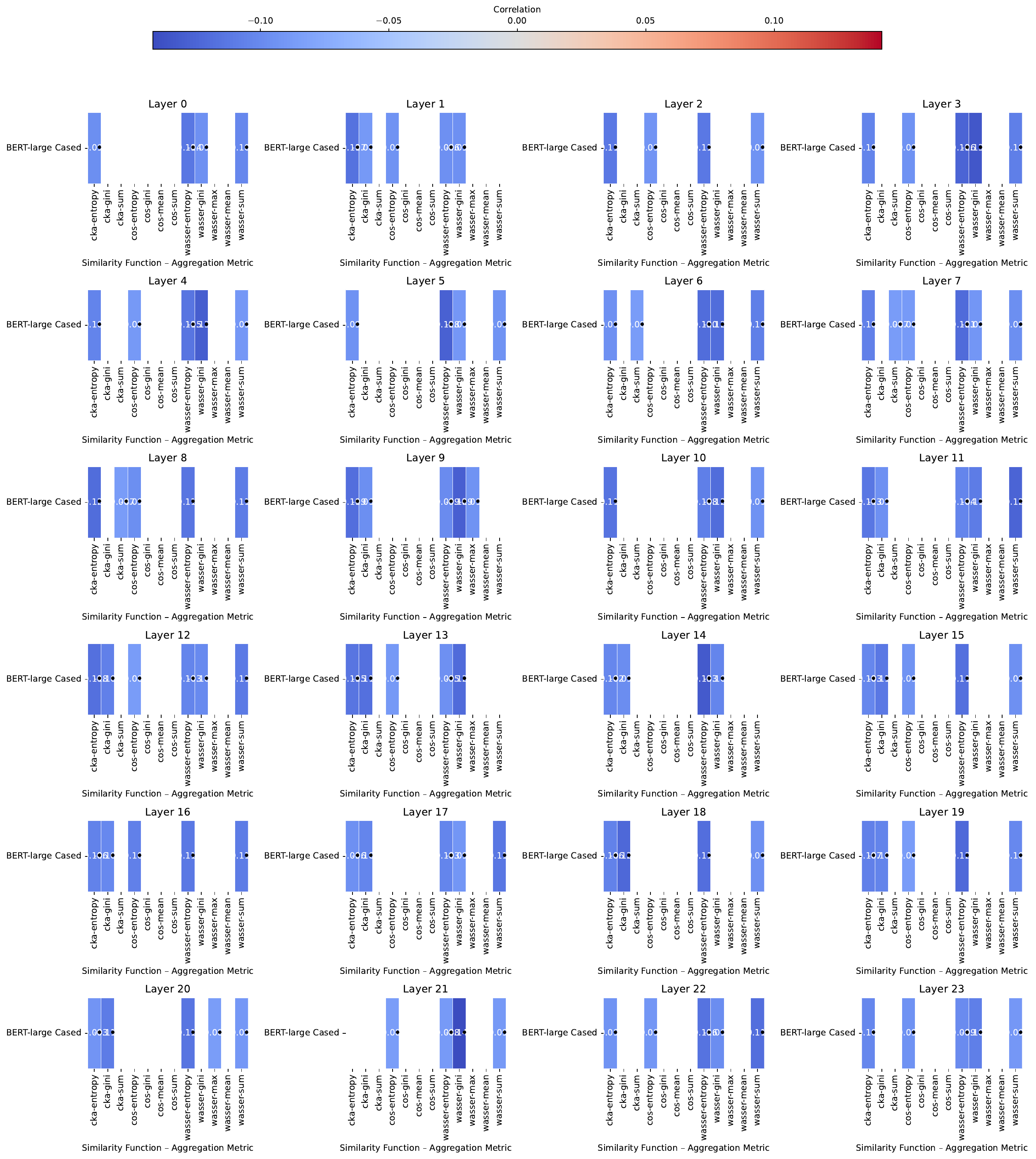}
    \caption{Correlation results for BERT-large Cased}
    \vspace{-5pt}
    \label{fig:correlation_bert_large_cased}
\end{figure*}

\begin{figure*}[t]
    \centering
    \includegraphics[width=0.9\textwidth]{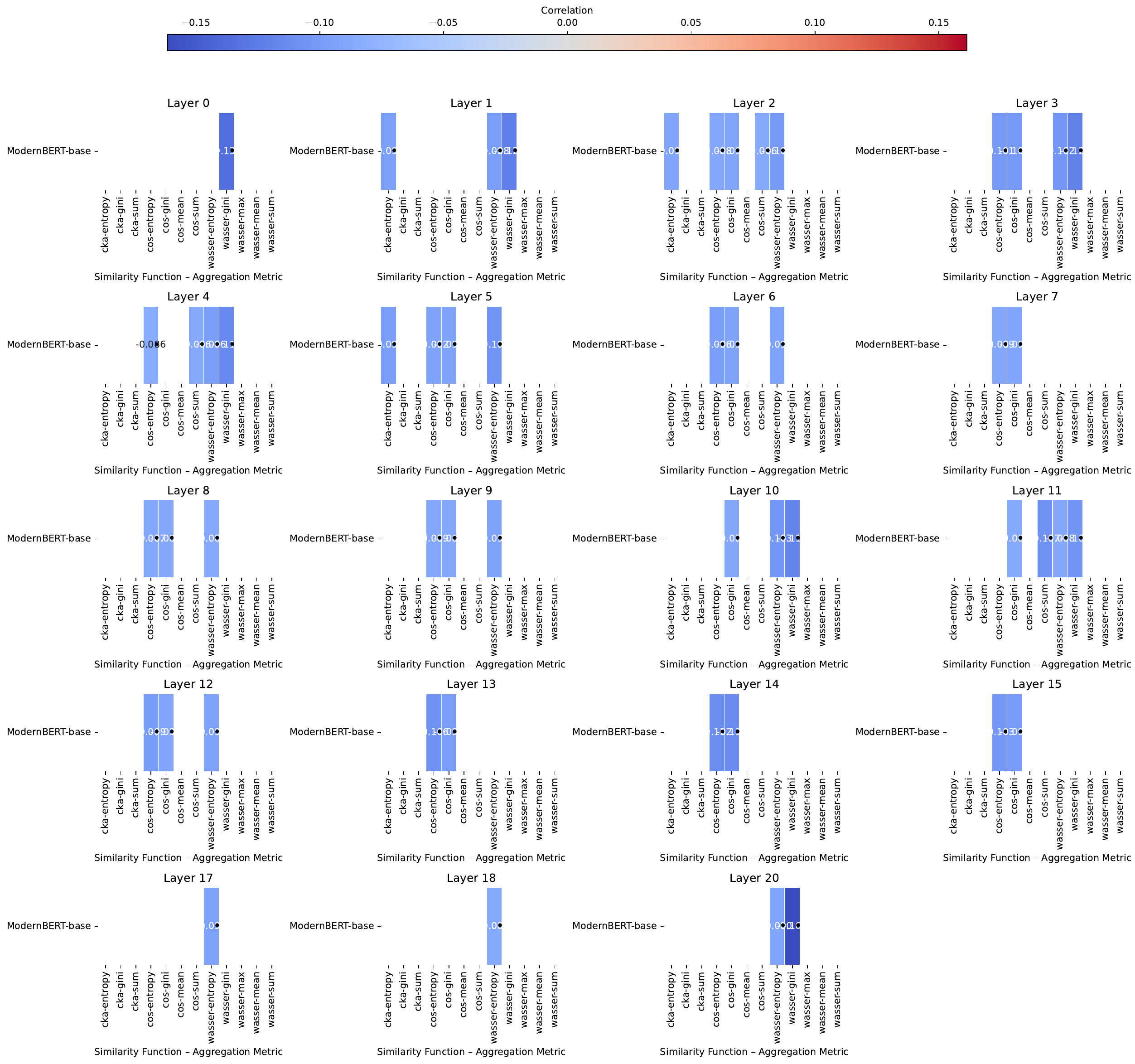}
    \caption{Correlation results for ModernBERT Base}
    \vspace{-5pt}
    \label{fig:correlation_modernbert_base}
\end{figure*}

\begin{figure*}[t]
    \centering
    \includegraphics[width=0.9\textwidth]{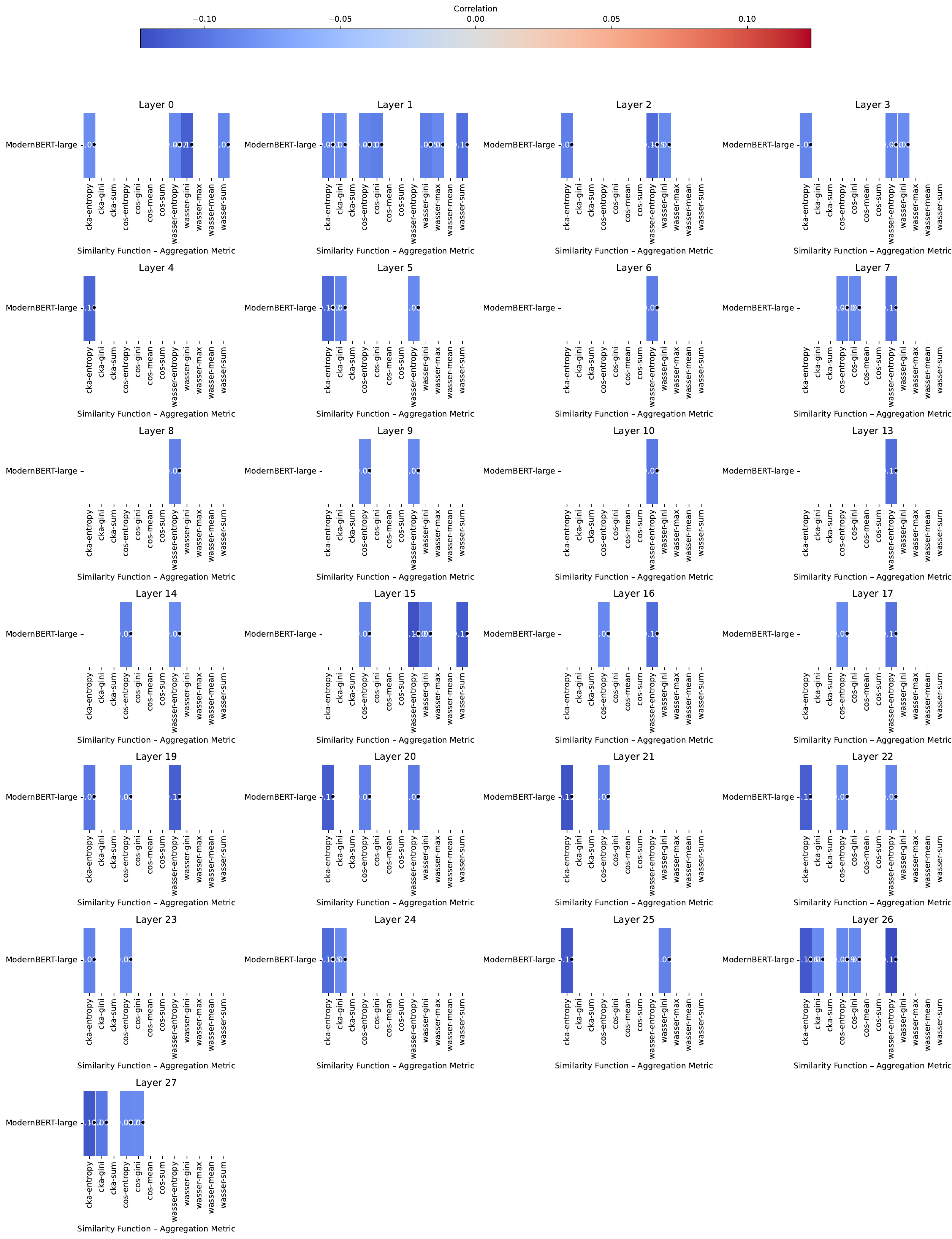}
    \caption{Correlation results for ModernBERT Large}
    \vspace{-5pt}
    \label{fig:correlation_modernbert_large}
\end{figure*}

We present the Spearman ranked correlation results of layer-wise decomposability derivations and syntactic flexibility in the six bidirectional models we have tested.
Empty bars show non-statistically significant results. For each model, the results are presented in its corresponding figure:

\begin{itemize}
    \item BERT-base (Uncased): \Cref{fig:correlation_bert_base_uncased}
    \item BERT-base (Cased): \Cref{fig:correlation_bert_base_cased}
    \item BERT-large (Uncased): \Cref{fig:correlation_bert_large_uncased}
    \item BERT-large (Cased): \Cref{fig:correlation_bert_large_cased}
    \item ModernBERT-base: \Cref{fig:correlation_modernbert_base}
    \item ModernBERT-large: \Cref{fig:correlation_modernbert_large}
\end{itemize}

\subsubsection{Linear Regression Results}
\label{sec:linear_regression_interaction_full_results}

Table~\ref{tab:regression_full} reports the full set of coefficient estimates from an ordinary least squares (OLS) regression predicting \textit{score}. The model includes fixed effects for network layer (treated as a categorical variable) and model identity, as well as several standardised continuous predictors and their interactions.

All continuous predictors (\texttt{steps}, \texttt{log\_frequency}, \texttt{surprisal}, and \texttt{decomp}) were z-standardised prior to estimation. Interaction terms capture moderation of each linguistic predictor by \texttt{steps}. The model was estimated using heteroskedasticity-robust (HC3) standard errors.

The table reports coefficient estimates, robust standard errors, z-statistics, two-sided p-values, and 95\% confidence intervals.

\section{Frequency Extraction with Infini-Gram}
\label{sec:unlemmatisation}
Whilst Infini-gram can approximate the frequency of expressions in model's training data, Infini-Gram does not support lemma-based queries -only exact strings. Thus, we devise a method to "unlemmatise" the expressions to try and capture as many variants as possible. 
Consequently, we convert idioms from their base form to surface forms in three steps:
\begin{enumerate}
    \item Add all verb inflections
    \item Add both singular and plural noun forms
    \item Replace "somebody" and "something" with appropriate possessive pronouns or possessive adjectives, depending on their syntactic position.
\end{enumerate}

We use the \texttt{word\_forms} package to obtain singular and plural noun forms, as well as all morphological variants of verbs.
\paragraph{Example.} Idiom: \textit{break somebody's heart}
\begin{enumerate}
    \item break: break, breaking, broke, broken
    \item somebody's: my, your, his, her, its, our, their
    \item heart: heart, hearts
\end{enumerate}
For this idiom, we therefore produce (4 x 7 x 2) 56 distinct queries.

\begin{table*}[!htp]\centering
\begin{tabular}{lrrrrrrr}\toprule
&coef &std err &z &P>|z| &[0.025] &[0.975] \\\midrule
Intercept &0.9142 &0 &3102.648 &0 &0.914 &0.915 \\
C(layer)[T.1] &0.0425 &0 &123.268 &0 &0.042 &0.043 \\
C(layer)[T.2] &0.0423 &0 &124.69 &0 &0.042 &0.043 \\
C(layer)[T.3] &0.0354 &0 &102.021 &0 &0.035 &0.036 \\
C(layer)[T.4] &0.0354 &0 &102.53 &0 &0.035 &0.036 \\
C(layer)[T.5] &0.0346 &0 &100.533 &0 &0.034 &0.035 \\
C(layer)[T.6] &0.0385 &0 &114.276 &0 &0.038 &0.039 \\
C(layer)[T.7] &0.0394 &0 &117.605 &0 &0.039 &0.04 \\
C(layer)[T.8] &0.0414 &0 &124.666 &0 &0.041 &0.042 \\
C(layer)[T.9] &0.0432 &0 &131.538 &0 &0.043 &0.044 \\
C(layer)[T.10] &0.0419 &0 &126.643 &0 &0.041 &0.043 \\
C(layer)[T.11] &0.0435 &0 &132.437 &0 &0.043 &0.044 \\
C(layer)[T.12] &0.0437 &0 &132.897 &0 &0.043 &0.044 \\
C(layer)[T.13] &0.0438 &0 &133.522 &0 &0.043 &0.044 \\
C(layer)[T.14] &0.0433 &0 &131.941 &0 &0.043 &0.044 \\
C(layer)[T.15] &0.0432 &0 &131.702 &0 &0.043 &0.044 \\
C(layer)[T.16] &0.0416 &0 &126.204 &0 &0.041 &0.042 \\
C(layer)[T.17] &0.0399 &0 &120.001 &0 &0.039 &0.041 \\
C(layer)[T.18] &0.0386 &0 &115.658 &0 &0.038 &0.039 \\
C(layer)[T.19] &0.038 &0 &113.435 &0 &0.037 &0.039 \\
C(layer)[T.20] &0.0363 &0 &107.521 &0 &0.036 &0.037 \\
C(layer)[T.21] &0.0358 &0 &105.709 &0 &0.035 &0.036 \\
C(layer)[T.22] &0.0342 &0 &99.947 &0 &0.033 &0.035 \\
C(layer)[T.23] &0.0335 &0 &97.523 &0 &0.033 &0.034 \\
C(layer)[T.24] &0.0336 &0 &97.999 &0 &0.033 &0.034 \\
C(layer)[T.25] &0.0334 &0 &97.374 &0 &0.033 &0.034 \\
C(layer)[T.26] &0.0337 &0 &98.104 &0 &0.033 &0.034 \\
C(layer)[T.27] &0.0344 &0 &100.512 &0 &0.034 &0.035 \\
C(layer)[T.28] &0.0366 &0 &107.535 &0 &0.036 &0.037 \\
C(layer)[T.29] &0.0414 &0 &123.712 &0 &0.041 &0.042 \\
C(layer)[T.30] &0.0537 &0 &169.44 &0 &0.053 &0.054 \\
C(layer)[T.31] &0.0648 &0 &212.713 &0 &0.064 &0.065 \\
C(layer)[T.32] &0.0653 &0 &215.456 &0 &0.065 &0.066 \\
C(model)[T.Olmo-3-1025-7B] &0.0041 &6.04e-5 &67.731 &0 &0.004 &0.004 \\
steps\_z &0.0037 &2.77e-5 &134.436 &0 &0.004 &0.004 \\
log\_frequency\_z &0.0085 &3.70e-5 &229.318 &0 &0.008 &0.009 \\
surprisal\_z &-0.0065 &3.21e-5 &-202.956 &0 &-0.007 &-0.006 \\
decomp\_z &0.0099 &2.93e-5 &336.962 &0 &0.01 &0.01 \\
steps\_z:log\_frequency\_z &-0.0008 &3.39e-5 &-24.692 &0 &-0.001 &-0.001 \\
steps\_z:surprisal\_z &-0.0007 &3.13e-5 &-22.301 &0 &-0.001 &-0.001 \\
steps\_z:decomp\_z &-0.001 &2.79e-5 &-36.367 &0 &-0.001 &-0.001 \\
\bottomrule
\end{tabular}
\caption{Full OLS regression results predicting \textit{score}. The model includes layer and model fixed effects, z-standardised predictors (steps, log frequency, surprisal, and decomposability), and interactions between steps and each linguistic predictor. Robust (HC3) standard errors are reported.}\label{tab:regression_full}
\end{table*}

\end{document}

%% file: sections/1-introductions.tex
Idiomatic expressions, such as ``\textit{spill the beans}'' (meaning to reveal a secret), ``\textit{pop the question}'' (propose marriage), or ``\textit{kick the bucket}'' (to die), exhibit a well-known tension between fixed form and non-compositional meaning. Although idiomatic meanings are always not fully predictable from the composition of the meanings of their individual parts \citep{sag_pain,fraser1970idioms,strassler1982idioms}, idioms vary in how transparently those parts relate to the overall interpretation \citep{nunberg1978pragmatics,nunberg1994idioms}. In some cases, individual words map transparently onto aspects of the figurative meaning (e.g., \textit{spill} $\rightarrow$ reveal, \textit{beans} $\rightarrow$ secrets). In others, the figurative meaning cannot be attributed to any constituent (e.g., \textit{kick, bucket,} $\rightarrow$ \textit{?}). This property, defined as \emph{decomposability} \citep{Gibbs1989, wasow1983idioms}, has been linked to various behavioural and grammatical phenomena, most notably patterns of syntactic flexibility \citep{nunberg1978pragmatics}.

This intuition is formalised in the Idiom Decomposability Hypothesis (IDH; \citep{gibbs_nayak_1989, Gibbs1989}). The IDH holds that idioms differ in semantic decomposability and that speakers share stable intuitions about how constituent meanings contribute to the whole. These intuitions are predicted to govern syntactic behaviour: the more decomposable an idiom is, the more syntactic flexibility it allows.

Decomposability has played a central role in prior works not merely as a descriptive notion, but as part of a broader explanatory account. Hybrid processing theories argue that decomposability reflects internal semantic structure \citep{CacciariLevorato1998}. According to this view, when an idiom’s figurative meaning is distributed over its constituent words, those constituents remain accessible to grammatical operations. Consequently, an idiom can undergo syntactic modification without losing its figurative interpretation \citep{nunberg1994idioms,gibbs_nayak_1989,Riehemann:2001}. This predicts a systematic relationship between semantic decomposability and syntactic flexibility, as illustrated in (1–4). While the decomposable idiom ``\textit{pop the question}'' retains its figurative interpretation under passivisation (1-2), the non-decomposable idiom ``\textit{kick the bucket}'' does not (3-4).

\lb{1}{He \textbf{popped the question}.}
\lb{2}{The \textbf{question} \textcolor{blue}{was} \textbf{popped} \textcolor{blue}{by him}.}
\lb{3}{He \textbf{kicked the bucket.}}
\lb{4}{\# The \textbf{bucket} \textcolor{blue}{was} \textbf{kicked} \textcolor{blue}{by him}.\footnote{\# is used to denote loss of idiomatic meaning.}}

An alternative perspective emerges from usage-based and constructionist approaches to linguistic knowledge. These accounts argue that form-meaning pairings are learned through exposure. Such pairings become entrenched as a function of frequency, predictability, and contextual diversity, rather than being derived from constituent-to-meaning mappings \citep{Bybee_2010,10.1093/acprof:oso/9780199268511.001.0001}. Applied to idioms, this perspective predicts that syntactic behaviour reflects distributional experience and familiarity. It also implies that the effects of decomposability may be unstable across tasks and items, as observed in psycholinguistic studies \citep{libben2008multidetermined,Tabossi2009,nordmann2014}. These two perspectives therefore differ in what they take to be the source of idiom generalisations: internal semantic structure on the one hand, or distributional experience on the other.

Human ratings of decomposability have long served as the primary empirical basis for evaluating these competing accounts \citep{partee_1995}. While such judgments provide valuable insight into speakers' linguistic intuitions, they reflect the full range of cognitive resources humans bring to language use and acquisition \citep{hubbard2023predictability}. As a result, they offer limited leverage on a distinct but fundamental question: which aspects of idiom behaviour arise from exposure to usage alone? Contextualised language models provide a principled way to probe this question. Trained on large amounts of text, these models acquire representations through distributional exposure, without explicit access to semantic role structure or acceptability judgments \citep{chang_word_learning,openai2022chatgpt,llama,ficarra-etal-2025-distributional}. Studying idioms in this setting isolates what can be learned from distributional exposure alone \citep{mi-etal-2025-rolling}.



In this work, we use contextualised language models as controlled distributional learners to investigate the relationship between decomposability, syntactic flexibility, and usage-based factors. We introduce a representational diagnostic of decomposability and relate it to human judgments, corpus-based measures of syntactic flexibility, and predictability. This perspective allows us to revisit the classic IDH theory from a mechanism-sensitive standpoint. Together, these analyses test whether decomposability explains idiom learning under purely distributional exposure, or whether usage-based factors account for the same variance.
We address the following \textbf{research questions}:
\begin{enumerate}
    \item To what extent do measures of decomposability derived from models align with decomposability judgments by humans? (\Cref{sec:human-ratings-correlation})
    \item How does representational decomposability relate to syntactic flexibility and usage-based factors such as predictability in a distributional learner? (\Cref{sec:ratings_composition})
    \item How do idiomatic representations evolve during pretraining, and are these dynamics better explained by decomposability or by usage-based predictability? (\Cref{sec:aot})

\end{enumerate}

\paragraph{Contributions.}
We make three contributions. First, we show that model-derived decomposability exhibits only a weak positive correspondence with human judgments. This suggests that the two operationalisations reflect overlapping but distinct idiom properties. Second, we find a weak but consistent negative correlation between representational decomposability and corpus-based measures of syntactic flexibility across models and layers, with the strongest effects for prepositional phrasal idioms, contrary to decomposability-based predictions. Third, analysing training checkpoints reveals that the emergence and stabilisation of idiomatic representations are best explained by surprisal and decomposability. Frequency alone offers limited explanatory contribution, underscoring the dominant role of distributional predictability over raw exposure.

%% file: sections/2-related_work.tex
\paragraph{IDH and the Compositionality Debate.}
Early accounts treated phrasal idioms as noncompositional units whose meanings are not derived from their parts \citep{Chomsky_1980, fraser1970idioms,heringer, Katz1973}. While this view captures the semantic opacity of many idioms, it offers limited explanation on why idioms differ systematically in their syntactic flexibility. 
In response, later work proposed the Idiom Decomposability Hypothesis (IDH), according to which idioms vary in the extent to which their figurative meanings can be distributed over their component words, and this variation constrains grammatical behaviour. In influential psycholinguistic work, \citet{gibbs_nayak_1989} show that many idioms are judged to be partially decomposable, and that speakers’ beliefs about decomposability predict acceptability judgments of syntactic alternations. On this view, differences in idiom flexibility are taken to reflect internal semantic structure, shaped by semantic and pragmatic factors in addition to syntax \citep{nunberg1994idioms}. At the same time, subsequent work has shown that decomposability judgments are gradient, task-dependent, and subject to considerable inter-speaker variability, and that their relationship to syntactic flexibility is less robust than originally assumed \citep{libben2008multidetermined,Tabossi2009, wierzba2023sources, sheinfux2019verbal}.
This empirical tension motivates our work, revisiting the decomposability-flexibility relationship from the perspective of distributional learning.

\paragraph{Decomposability and Idiom Processing.}
A related line of work has examined whether idiom decomposability influences online processing and recognition. Early accounts propose that decomposable idioms are processed compositionally, using the same lexical retrieval and syntactic parsing mechanisms as literal language, whereas non-decomposable idioms are retrieved as single lexical units \citep{gibbs1989speakers}. However, subsequent studies have challenged this sharp distinction \citep{sprenger2006lexical, Tabossi2009Idioms}. In particular, \citet{Tabossi2009Idioms} find no interaction between decomposability and processing speed in semantic judgment tasks, with both idiom types recognised equally quickly. More broadly, these findings suggest that decomposability effects are task-dependent and may not generalise across different aspects of idiom behaviour, such as recognition speed versus syntactic flexibility. This variability further motivates a mechanism-sensitive examination of which idiom patterns emerge under different learning and processing settings.
 


\paragraph{Challenges with Quantifying Decomposability.}
Human ratings have been central to investigations of IDH. However, several norming and processing studies suggest that such judgments are neither uniform nor stable across speakers and tasks.
A norming study by \citet{Titone_norms} found that fewer than half of idioms elicited consistent judgments, with substantial inter-speaker disagreement. This pattern persisted in later eye-tracking work \citep{Titone1999}, which revealed only weak, participant-specific effects of decomposability. As observed by \citet{libben2008multidetermined}, idioms are ``\textit{multidetermined}'' and \citet{Tabossi2009} suggests that, these findings undermine the assumption that idioms can be cleanly classified as decomposable or not based on shared intuitions.
These factors motivate our work, which complements this line of research by examining decomposability from a perspective that does not rely solely on human semantic judgments.

%% file: sections/3-formulisations.tex
\subsection{Theoretical Stipulations}
We formalise decomposability under a set of theoretical stipulations grounded in prior work:

\begin{itemize}
    \item \textbf{(IDH) Decomposability-syntactic flexibility link:} Decomposability is assumed to correlate (positively) with syntactic flexibility \citep{nunberg1994idioms}.
    
    \item \textbf{(S1) Semantic alignment:} Decomposability can be understood as the degree of semantic alignment between constituent representations and the idiom’s overall figurative meaning \citep{nunberg1994idioms}.
    
    \item \textbf{(S2) Distributed meaning:} Decomposability reflects the extent to which an idiom’s figurative meaning is distributed across its constituent parts \citep{nunberg1994idioms}.
    
    \item \textbf{(S3) Gradient:} Decomposability and syntactic flexibility are gradient properties that vary along a continuum\footnote{Earlier accounts have proposed categorical distinctions, including a binary classification \citep{nunberg1994idioms} a three-way distinction \citep{gibbs_nayak_1989}, whereas more recent work argues for a continuum-based view \citep{sheinfux2019verbal}.}\citep{sheinfux2019verbal}.
\end{itemize}

By operationalising decomposability in a manner consistent with \textbf{(S1--S3)}, our formulation allows us to test the Idiom Decomposability Hypothesis (\textbf{IDH}) independently, by evaluating whether the resulting decomposability measure predicts syntactic flexibility in usage.

\subsection{Decomposability} \label{subsec:decomp}

Idioms are defined as contextually dependent semantic units characterised by a tension between their holistic, figurative interpretation and the literal meanings of their constituent words \citep{gibbs2012interpreting,nunberg1994idioms}. Figurative meanings arises when an expression’s interpretation cannot be compositionally derived from its parts and instead emerges in context. Modelling this form of non-compositionality therefore requires representations integrating information across the entire expression while also providing context-sensitive token-level representations. 
Contextualised language models meet these requirements by jointly encoding sentence-level context and token-level representations learned from distributional exposure \citep{vaswani_attention,devlin-etal-2019-bert}. In particular, bidirectional transformer-based models  allow each token representation to be informed by its surrounding context, making them well-suited for analysing how figurative meaning is distributed across idioms.


From this perspective, hidden-state geometry offers a principled basis for operationalising decomposability. If an idiom is decomposable, we expect that (i) perturbing the idiom's constituent tokens will induce systematic changes in the hidden-state representation of the full expression, and (ii) the influence of individual constituents will be reflected in, and partially traceable from, the resulting representations.
Conversely, if an idiom is non-decomposable, modifying a component should have weaker or less systematic effects on the model's internal representation of the sentence.




Let $f$ denote a pretrained bidirectional transformer encoder (BERT) that maps an input sentence to contextualised token representations. For an input sentence $s = (w_1,\dots,w_n)$, the encoder produces hidden states
\[
\mathbf{h}_j(s) = f(s)_j \in \mathbb{R}^d, \qquad j = 1,\dots,n.
\]
where $j$ denotes the token position. Because the encoder is bidirectional, each token representation incorporates information from both left and right contexts. This is important for modelling idioms as multi-token expressions whose figurative meaning depends on context.

We obtain a sentence-level representation via a pooling operation $P(\cdot)$ over token representations:
\begin{equation}
\mathbf{e}(s) = 
P\!\left( \mathbf{h}_1(s), \dots, \mathbf{h}_n(s) \right). \nonumber
\end{equation}

Let $s_g$ denote the sentence obtained by replacing the idiom in context with a paraphrastic figurative gloss expressing its intended meaning. The corresponding pooled representation is:
\begin{equation}
\mathbf{e}(s_g) = 
P\!\left( \mathbf{h}_1(s_g), \dots, \mathbf{h}_{n_g}(s_g) \right). \nonumber
\end{equation}


In line with \textbf{S1}, we operationalise semantic alignment between an idiom in context and its figurative meaning. Let $\mathrm{sim}(\cdot,\cdot)$ denote a similarity measure between sentence representations (e.g., cosine similarity, Centered Kernel Alignment (CKA) \citep{kornblith2019similarity}, or the Wasserstein distance \citep{Kantorovitch1958,Vaserstein1969Markov}). We define the figurative similarity score for the full idiom as $S_{\mathrm{fig}} = \mathrm{sim}\!\left(\mathbf{e}(s), \mathbf{e}(s_g)\right).$

To estimate token-level contributions, let $I \subseteq \{1,\dots,n\}$ denote the set of token indices corresponding to the idiom span. For each idiom token $j \in I$, we construct a leave-one-out variant 
by masking that token:
\begin{equation}
s^{(-j)} = (w_1,\dots, w_{j-1}, \texttt{[MASK]}, w_{j+1},\dots, w_n). \nonumber
\end{equation}

We compute the pooled representation $\mathbf{e}(s^{(-j)})$ and its similarity to the gloss-replaced sentence:
\begin{equation}
S_{\mathrm{mask}}^{(j)} =
\mathrm{sim}\!\left( \mathbf{e}\!\left(s^{(-j)}\right),\, \mathbf{e}(s_g) \right). \nonumber
\end{equation}

If a constituent contributes to the idiomatic meaning, removing it should disrupt the alignment between the idiom-in-context sentence and its figurative paraphrase. We therefore define the contribution of token $j$ to the idiomatic meaning as the absolute change in similarity when it is masked, consistent with the semantic alignment assumption in \textbf{S2}\footnote{See \Cref{sec:decomp_token_example} for worked examples}. Taking the absolute value ensures contributions are non-negative; using signed values would allow large positive and negative effects to cancel, thus obscuring each token’s true influence and rendering distributional measures unstable or uninterpretable.
\begin{equation}
\Delta_j = \left|S_{\mathrm{fig}} - S_{\mathrm{mask}}^{(j)}\right|. \nonumber
\end{equation}

\paragraph{Aggregation.}

Given the set of token-level contributions scores for an idiom, $
\mathcal{D} = \{ \Delta_j \mid j \in I \}$,
 we obtain an expression-level decomposability score by aggregating these values. Aggregation is required because decomposability is defined at the level of the idiomatic expression, while contributions are estimated at the level of individual constituents. We consider four aggregation functions that capture different aspects of contribution distribution across constituents, including mean, maximum, Gini dispersion, and entropy\footnote{Formal definitions of these aggregation functions are provided in \Cref{sec:agg_func}.}. This produces a continuous measure of decomposability, consistent with \textbf{S3}.








\subsection{Syntactic Flexibility} \label{subsec:flexibility}

We operationalise syntactic flexibility as the diversity of constructional environments in which an idiom occurs. Specifically, we group each attested occurrence of an idiom by constructional type and quantify flexibility as the evenness of its distribution across these types. Following \citet{Tabossi2009} and \citet{gibbs_nayak_1989}, we distinguish four constructional types in addition to the base form: adverb insertion, adjective insertion, passivization, and action nominalization. Idioms that are highly flexible occur across multiple constructions, whereas rigid idioms are concentrated in a single form.

Let $C$ denote the set of mutually exclusive construction types covering the attested syntactic realisations of an idiom. For an idiom $i$, we estimate the probability of construction $c \in C$ from corpus counts as $p_{i,c} = \frac{n_{i,c}}{N_i}$, where $n_{i,c}$ is the number of occurrences of idiom $i$ in construction $c$, and $N_i = \sum_{c \in C} n_{i,c}$ is the total number of occurrences of idiom $i$.
We define the syntactic flexibility of idiom $i$ as the Shannon entropy of its constructional distribution, which captures both the diversity of constructions an idiom occurs in and how evenly its occurrences are distributed across them. 

\begin{equation}
    H(i) = -\sum_{c \in C}  p_{i,c} \log_2 p_{i,c}. \nonumber
\label{eq:entropy_syntactic_flexibility}
\end{equation}
Entropy increases as an idiom appears in a larger number of constructions and as its occurrences are distributed more evenly across them. The maximum entropy occurs when all construction types are equally probable:
    $H_{max} = \log_2(|C|)$.
We provide a worked example in \Cref{syntactic flexibility example}.

\subsection{Frequency} \label{subsec:frequency}
The frequency of an idiom is calculated as the sum of its counts across all constructional frames described above.

\subsection{Predictability} \label{subsec:predictability}

Following prior idiom norming work, we operationalise predictability as the probability of the idiom-final word given its preceding context (i.e., cloze completion; e.g., \citet{Vulchanova2019Boon,CACCIARI1988}). This captures the degree of contextual constraint: once sufficient context is processed, the final word often completes the idiomatic configuration and triggers the figurative interpretation (Configuration Hypothesis; \citep{cacciari_1991,Titone_norms}). As such, predictability reflects distributional expectations learned from usage rather than semantic transparency alone. To remain comparable with human norms, we use the model analogue, $\log P(\text{final word} \mid \text{context})$, computed via masked prediction in bidirectional models. When the final word is split into multiple subword tokens, we calculate predictability as the average of their log probabilities.







%% file: sections/4_experiments.tex
\subsection{Models}
We employ bidirectional transformer models, as idiomatic interpretation often depends on information from both preceding and following context. Psycholinguistic evidence from eye-tracking studies indicates that human readers consult surrounding context during idiom interpretation \citep{Titone1999}; we treat this evidence as a motivation for using architectures that explicitly encode both left and right context. Accordingly, we use bidirectional architectures, namely, BERT \citep{devlin-etal-2019-bert} \texttt{base, large} variants and ModernBERT \citep{modernBERT}, a reimplementation of the BERT architecture with updated training practices, which conditions on context of both sides of an idiom. Details for the models are provided in \Cref{sec:model_info}.

\subsection{Datasets}
We use datasets composed of sentences in which idiomatic expressions are employed figuratively (e.g., “How have you \textit{weathered the storm}?”), paired with glossed paraphrases that convey the same meaning without the idiom (e.g., “How have you \textit{succeeded in getting through the difficult situation}?”). Our experiments draw on the IMPLI dataset \citep{stowe2022impli}, which contains figurative sentences together with paraphrases in which each idiom is replaced by the gloss. A summary of the datasets is provided in \Cref{sec:app_Datasets}.

\subsection{Frequency Extraction}
We extract idiom frequencies from the enTenTen corpus via Sketch Engine \citep{ententen}\footnote{\texttt{www.sketchengine.eu}}. enTenTen is a large, web-scale corpus of English that provides broad coverage of contemporary usage and supports lemmatised queries, making it well suited for estimating idiom frequencies across diverse syntactic contexts. For each idiom, we lemmatise its base form and generate Corpus Query Language (CQL) patterns corresponding to each syntactic frame. We then query the corpus and record the resulting frequency counts.

\subsection{Evaluation of Hypotheses}
We evaluate competing accounts of idiom decomposability through a set of complementary analyses examining the relationship between representational decomposability, syntactic flexibility, and usage-based factors. 
In line with the IDH, we assess whether decomposability is positively associated with syntactic flexibility by computing Spearman’s rank correlations between decomposability scores and entropy-based flexibility measures. To test usage-based accounts, we additionally examine how representational decomposability relates to frequency and predictability, and whether these factors account for variance attributed to decomposability in human judgments. 
All correlations are computed for representations extracted from each layer of each model, allowing us to assess the consistency of effects across representational levels. 
Finally, we analyse how idiomatic representations evolve over training time, testing whether decomposability or usage-based predictability better explains stabilisation dynamics during pretraining. 
We compare different similarity functions and embedding aggregation strategies to ensure that observed patterns are not driven by a particular representational choice.



%% file: sections/5_results.tex
This section assesses the relationship between idiom decomposability and syntactic flexibility across human judgments and model-derived representations. We first test the classic decomposability-flexibility claim using human ratings, then examine how decomposability is encoded in contextualised language models, and finally assess the role of usage-based factors. Given the large number of model configurations, layers, and decomposability metrics, we focus on robust qualitative patterns, full results are provided in \Cref{sec:layer-wise-corr-syntax}.

\subsection{Human Decomposability Ratings and Syntactic Flexibility}

We first test a central assumption of the Idiom Decomposability Hypothesis: that idioms judged as more decomposable are more syntactically flexible \citep{nunberg1994idioms}. Contrary to this assumption, we find no significant relationship between human decomposability ratings and corpus-based measures of syntactic flexibility across the 90 idioms shared between the Bulkes and Tanner dataset and IMPLI. This result aligns with prior work questioning the stability and predictive power of decomposability judgments \citep{dolev2025decomposability} and suggests that, while such ratings reflect perceived semantic transparency, they do not reliably predict syntactic behaviour. These findings motivate a broader comparison: if decomposability judgments are poor predictors of syntactic behaviour in human data, how do they relate to decomposability as encoded in distributional models?

\begin{figure}[h]
\centering
\begin{adjustbox}{width=1\columnwidth}
\includegraphics{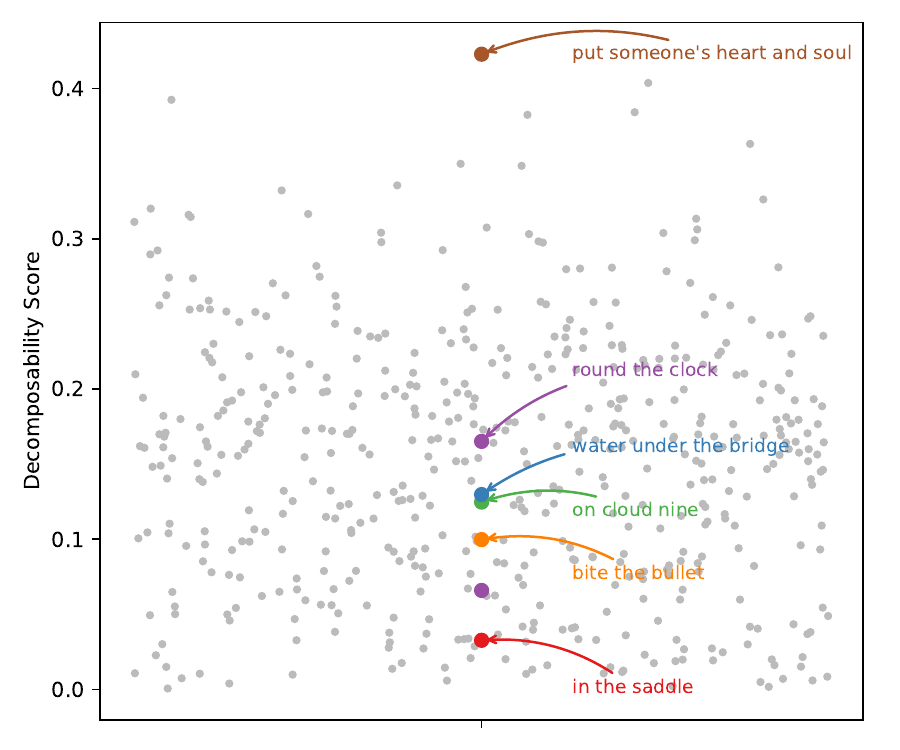}
\end{adjustbox}
  \caption{Decomposability scores obtained from the set-up that most resembles human decomposability ratings. Most phrases cluster in the low-to-mid range, while ``\textit{put someone’s heart and soul}'' stands out as highly decomposable and ``\textit{in the saddle}'' as among the least.}
  \label{fig:most_aligned_to_human_decomp}
\end{figure}

\subsection{Human Ratings and Model-Derived Decomposability}
\label{sec:human-ratings-correlation}
We use contextualised models to assess whether a distributional learner recovers patterns similar to those reflected in human decomposability judgments. Across models, layers, and representational configurations, correlations between human ratings and model-derived decomposability are consistently weak but positive. The strongest correspondence is observed for BERT-large (Uncased) using final-layer representations, Wasserstein distance, and sum-based aggregation ($r(90)=.24$, $p=.005$). \Cref{fig:most_aligned_to_human_decomp} illustrates the decomposability scores calculated in this way.

Although statistically reliable, with robustness checks reported in \Cref{sec:robust}, this effect size indicates only a partial overlap. These results suggest that contextualised language models encode some aspects of the intuitions underlying human decomposability ratings, while also diverging from them in systematic ways.

\subsection{Representational Decomposability and Syntactic Flexibility}
We next examine whether model-derived decomposability covaries with syntactic flexibility, as predicted by semantic accounts of idioms. Across models and layers, Spearman correlations are consistently small and frequently negative (maximum $r(527)=-.16$, $p=.0002$), indicating little systematic alignment between representational decomposability and syntactic variability.
Larger models, such as BERT-large and ModernBERT-large, yield a greater number of statistically significant correlations, though effect sizes remain modest. These effects are most consistently observed in earlier layers, particularly layer~2, which prior work has associated with local syntactic and phrasal information rather than abstract semantic composition \citep{jawahar-etal-2019-bert,tenney-etal-2019-bert}. 

\subsection{Idiom-Type Differences in the Decomposability-Flexibility Relation}

The relationship between decomposability and syntactic flexibility is not uniform across idiom types. Using the most human-aligned decomposability configuration, we analyse correlations separately by coarse syntactic category (\Cref{tab:pp_results_human}). 
Prepositional phrase idioms (PP+NP), such as ``\textit{off the hook}'' and ``\textit{in a nutshell}'', exhibit consistently stronger and statistically significant negative correlations than other idiom types. For these expressions, higher representational decomposability is associated with reduced syntactic flexibility.
This pattern plausibly reflects the morpho-syntactic rigidity of prepositional constructions: lacking a verbal head, they support fewer syntactic alternations, such as passivisation or inflection, independently of how figurative meaning is distributed across constituents. 
Notably, for verb phrase idioms-the class most directly targeted by the \textbf{IDH}-we do not observe a significant relationship between decomposability and syntactic flexibility. This absence of effect in the theoretically most relevant domain weakens support for a decomposability-based account.

\begin{table}[!htp]\centering
\resizebox{0.8\columnwidth}{!}{ 
\footnotesize
\begin{tabular}{lrrrr}\toprule
Structures &$n$ & $\rho$ &$p$ \\\midrule
VP &284 &-0.02 &0.68 \\
PP &127 &-0.24 &0.01* \\
NP &85 &-0.15 &0.18 \\\midrule
ADJP &15 &0.23 &0.40 \\
ADVP &10 &0.12 &0.75 \\
OTHER(NUM) &3 &- &- \\
S &2 &- &- \\
OTHER(INTJ) &1 &- &- \\
\bottomrule
\end{tabular}%
}
\caption{Spearman’s rank-order correlations between decomposability and syntactic flexibility by idiom type. Correlations are computed using decomposability measures derived from the most human-aligned setting (i.e., BERT-large uncased, last layer, Wasserstein and sum-based aggregreation); $n$ indicates the number of idioms per coarse syntactic category. * denotes significant results.}\label{tab:pp_results_human}
\end{table}

\subsection{Usage-Based Influences on Decomposability}
\label{sec:ratings_composition}

\begin{table}[!t]
\centering
\small
\begin{tabular}{l r r l}
\toprule
\multicolumn{4}{c}{\textbf{Human Ratings}} \\
\midrule
Variable & Coef & $z$ & $p$ \\
\midrule
Predictability & -0.52 & -0.33 & 0.73 \\
Frequency & -0.20 & -2.26 & 0.02* \\
Predictability x frequency & 0.03 & 0.24 & 0.80 \\
\midrule
\multicolumn{4}{c}{\textbf{Model-derived Measures (BERT Large (Cased)}} \\
\midrule
Variable & Coef & $z$ & $p$ \\
\midrule
Predictability & 0.002 & 1.43 & 0.15 \\
Frequency & -0.29 & -4.07 & 0.000* \\
Predictability x frequency & 0.002 & 0.98 & 0.32 \\
\bottomrule
\end{tabular}

\caption{Regression results for measures from humans and BERT Large (Cased). Frequency is log-transformed. * denotes significant results.}
\label{tab:results_human_models_rq2}
\end{table}

To clarify what decomposability ratings reflect, we examine their relationship with two usage-based variables: predictability and frequency. Frequency corresponds to corpus-derived usage statistics, while familiarity ratings from \citet{Bulkes2017} reflect perceived exposure; for more information on human-derived measures and their model-derived counterparts, see \Cref{tab:measures-human-vs-model}.

\begin{figure*}[tbp]
    \centering

    \begin{subfigure}{0.33\textwidth}
        \centering
        \includegraphics[width=\linewidth]{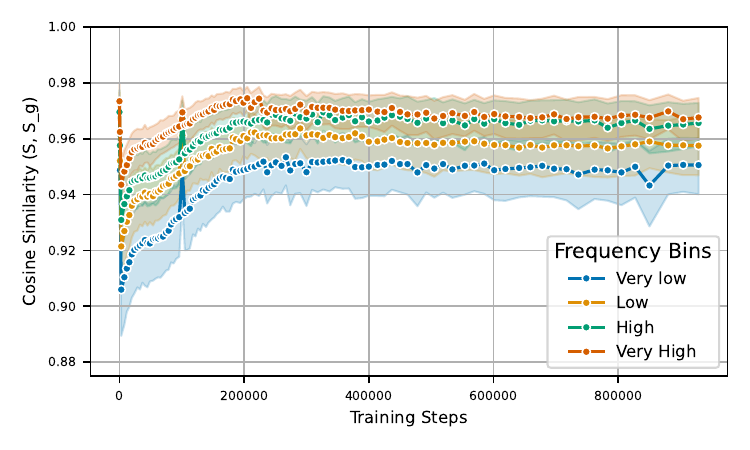}
    \end{subfigure}\hfill
    \begin{subfigure}{0.33\textwidth}
        \centering
        \includegraphics[width=\linewidth]{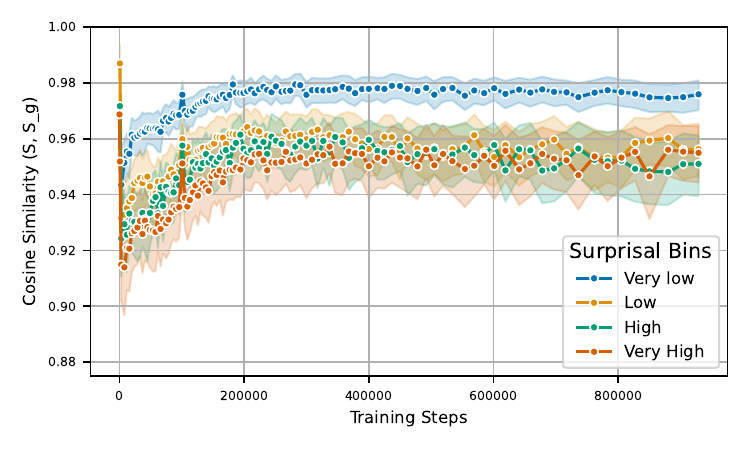}
    \end{subfigure}\hfill
    \begin{subfigure}{0.33\textwidth}
        \centering
        \includegraphics[width=\linewidth]{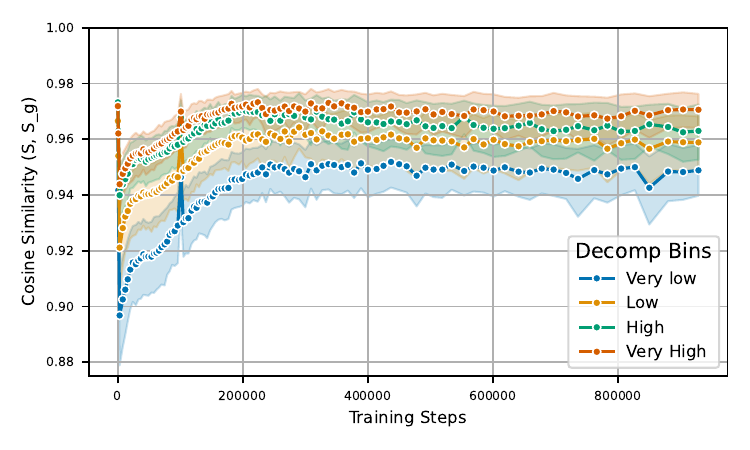}
    \end{subfigure}

    \medskip

    \begin{subfigure}{0.33\textwidth}
        \centering
        \includegraphics[width=\linewidth]{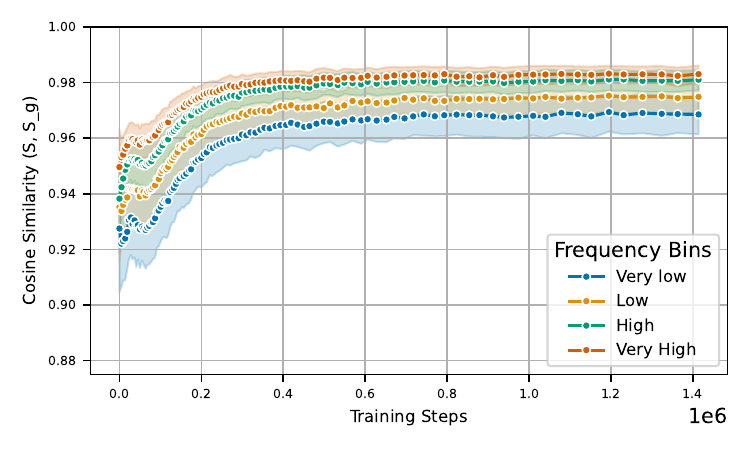}
        \caption{Frequency}
    \end{subfigure}\hfill
    \begin{subfigure}{0.33\textwidth}
        \centering
        \includegraphics[width=\linewidth]{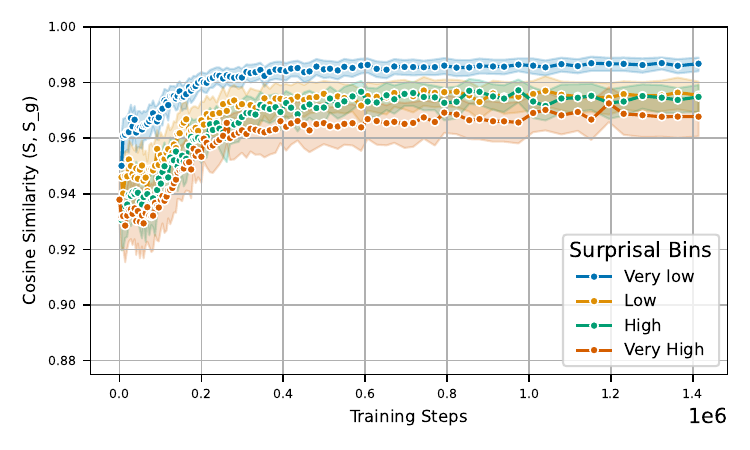}
        \caption{Surprisal}
    \end{subfigure}\hfill
    \begin{subfigure}{0.33\textwidth}
        \centering
        \includegraphics[width=\linewidth]{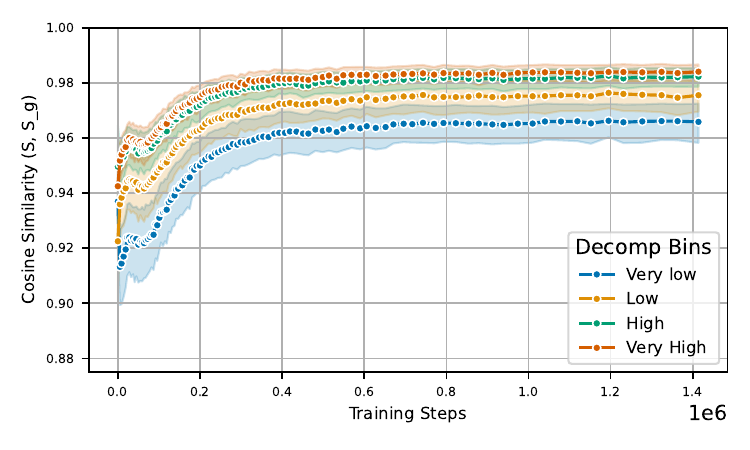}
        \caption{Decomposability}
    \end{subfigure}

    \caption{Representation similarity over pretraining for OLMo-2 7B (top row) and OLMo-3 7B (bottom row), measured across 100 checkpoints. Each checkpoint is plotted individually, with colour indicating idiom characteristics (frequency, surprisal, and decomposability). All results are from layer 13.}
    \label{fig:6subplots}
\end{figure*}

We fit separate regression models for human- and model-derived decomposability measures (results in \Cref{tab:results_human_models_rq2}). For human ratings, neither familiarity nor predictability yields a significant effect. However, familiarity and corpus frequency may index different constructs: familiarity reflects subjective experience, whereas frequency captures objective distributional regularities. When familiarity ratings are replaced with corpus-derived frequency counts, we observe a significant negative relationship between log-transformed frequency and decomposability ratings (coef $=-0.20$, $z=-2.26$, $p=0.02$), indicating that more frequent idioms are judged as less decomposable.
A parallel pattern emerges for model-derived decomposability. For BERT-large (cased), frequency shows a significant negative effect (coef $=-0.29$, $z=-4.07$, $p<.001$). No significant frequency effects are observed for other models.
The negative effect of frequency on decomposability aligns with previous findings that higher frequency lowers the processing cost of multi-word expressions \citep{arnon2010more, sosa2002evidence}. It suggests that frequently-used idioms are treated as single, holistic units that are stored and retrieved from memory, which is a characteristic of how non-decomposable idioms are processed \citep{Gibbs1989}.

%% file: sections/aot.tex
We examine how representations of idioms develop and become stable during pretraining in OLMo-2 (7B) \citep{olmo2} and OLMo-3 (7B) \citep{olmo3}, using internal representations extracted from 100 checkpoints for each model. This allows us to study idiom learning as a dynamic process rather than as a static end-state. Our analysis focuses on how idiom-intrinsic properties interact with usage-based predictability during representation formation.
 
Specifically, we track learning trajectories for idioms varying in decomposability and contextual surprisal, where surprisal captures predictability beyond raw frequency (see \Cref{fig:6subplots}). This design allows us to disentangle the roles of structural interpretability (decomposability) and distributional expectations (surprisal) \citep{mi-etal-2025-input} in shaping idiomatic representations over time, and to assess whether frequency alone \citep{mi-etal-2025-rolling} suffices to explain their emergence under exposure-only learning.

\paragraph{Frequency in Pretraining Data.} 
We estimate idiom frequency in the models’ pretraining data using Infini-gram \citep{InfiniGram}, which allows us to approximate the distribution of each idiom in the pretraining data as accurately as possible. See \Cref{sec:unlemmatisation} for additional details. 

\paragraph{Surprisal.} Derived from Shannon’s information theory \citep{shannon-1948}, surprisal quantifies how unexpected a word is within its context \citep{hale-2001-probabilistic,Levy2008Expectation}. It is defined as the negative log-probability of the token given its preceding context: $\text{Surprisal}(t_i) = -\log_2 P(t_i \mid t_{<i})$.

\paragraph{Decomposability.} We use the model-derived decomposability score that produced the highest correlation with human ratings for this analysis.

\paragraph{Performance Measure.}
 To track idiom learning, we measure representational similarity between a sentence containing an idiomatic expression ($S$) and a corresponding glossed sentence ($S_g$) that explicitly encodes the idiom’s intended figurative meaning. We compute this similarity using cosine similarity across layers and across 100 training checkpoints\footnote{We focus on \texttt{Stage 1} pretraining, prior to instruction tuning (\texttt{Stage 2}), and DPO (\texttt{Stage 3}) in OLMo's training.}.

\subsection{Results}

\begin{table}[!t]
\resizebox{\columnwidth}{!}{
\small
\begin{tabular}{lrrrr}
\toprule
 & Coef & z & p \\
\midrule
Steps x Frequency & -0.0008 & -24.69 & <0.001 \\
Steps x Surprisal     & -0.0007 & -22.301 & <0.001 \\
Steps x Decomposability & -0.0010 & -36.367 & <0.001 \\
\bottomrule
\end{tabular}
}
\caption{Interaction effects between training progress and idiomaticity characteristics. Coefficients from a linear regression of model scores on training steps and input properties. Robust (HC3) standard errors; all variables z-scored. Frequency is log-transformed.}
\label{tab:linear_regression_interaction}
\end{table}

\paragraph{Training Dynamics of Idiom Properties.}
We examine how the influence of idiom properties changes over training using a linear regression with interactions between training progress (steps) and three standardized predictors: log frequency, surprisal, and decomposability. All variables were z-scored, and HC3 robust standard errors were used. Results are summarised in \Cref{tab:linear_regression_interaction}; full results are presented in \Cref{sec:linear_regression_interaction_full_results}.

All three interaction terms are negative and highly significant (p < .001), which indicates that the effects of frequency, surprisal, and decomposability systematically diminish as training progresses. Pearson correlation analysis among the regression predictors further indicates that multicollinearity is unlikely to affect the estimates (see \Cref{sec:robust_pearson_corr_acquisition}).

\paragraph{Relative Contributions of Idiom Properties.} Among the three predictors, decomposability shows the largest interaction magnitude, which indicates the strongest training-dependent shift. This suggests that decomposability plays a more prominent role in shaping representations early in training, relative to frequency and surprisal.

\paragraph{Early Versus Late Training Effects.}
The observed interactions do not imply that idiom properties are inherently detrimental to learning. Rather, their total effect depends on training stage: they exert stronger influence early in training, with their impact attenuating as representations stabilise. This pattern is consistent with decreasing reliance on expression-level cues as the model accumulates distributional evidence over training.

%% file: sections/6-discussion.tex
Decomposability has long been central to debates about whether idiom behaviour reflects internal semantic structure or emerges from usage and exposure. In this work, we revisited this question using contextualised language models as controlled distributional learners, allowing us to isolate what can be learned from distributional information alone.

Our results offer a clear challenge to IDH. We found that decomposability exhibits limited and unstable links to syntactic flexibility-most notably failing to hold even within verb phrase idioms, the core empirical domain where the IDH should, in theory, be most robust. This absence of effect suggests that decomposability is not a reliable mechanism-neutral explanation for how idioms behave syntactically.

Instead of syntactic links, we find a robust negative relationship between frequency and decomposability, supporting the view that high-frequency idioms are represented holistically. However, our analysis of pretraining dynamics reveals a more nuanced picture: frequency alone fails to explain how idiomatic representations emerge. Rather, surprisal and decomposability drive the stabilization of these meanings over time.

Ultimately, our work demonstrates that investigating the internal representations of language models provides a powerful method for adjudicating theoretical debates. By isolating the signal available from exposure alone, we can distinguish between properties inherent to semantic structure and those that emerge naturally from distributional learning-thereby helping to disentangle the confounded factors that often underlie human linguistic judgments.

%% file: sections/8-limitations.tex
We recognise that our proposed decomposability metric represents just one possible way of operationalising this phenomenon. Decomposability is a complex property of idioms and has not yet been empirically studied. As such, we aim to offer an initial exploration in this direction; however, we do not claim that our approach is the only valid one.

A limitation of our pretraining analysis is the reliance on decomposability scores derived from a specific architecture (BERT-large) to predict the dynamics of another (OLMo). While using model-internal measures as fixed predictors is conceptually analogous to using human ratings or corpus-based frequency, model-derived decomposability is not architecture-neutral. Given that different models exhibit varying correlations with human judgments and syntactic flexibility, these scores may carry architecture-specific biases. Ideally, decomposability would be computed directly from the model under study; however, the context-sensitive nature of idiomaticity requires a bidirectional model, which is not applicable to causal models like OLMo. Future work should explore architecture-agnostic measures to ensure broader generalisability across different model families.

Finally, our analysis is based on a set of English idioms and may not extend to other languages. Investigating how decomposability manifests in cross-lingual studies is an interesting avenue for future work.


\section*{Ethical Considerations}
AI-assisted writing and coding tools were used in compliance with the ACL Policy on the Use of AI Writing Assistance.

%% file: custom.bib
@article{arnon2010more,
  title={More than words: Frequency effects for multi-word phrases},
  author={Arnon, Inbal and Snider, Neal},
  journal={Journal of memory and language},
  volume={62},
  number={1},
  pages={67--82},
  year={2010},
  publisher={Elsevier}
}

@article{sosa2002evidence,
  title={Evidence for frequency-based constituents in the mental lexicon: Collocations involving the word of},
  author={Sosa, Anna Vogel and MacFarlane, James},
  journal={Brain and language},
  volume={83},
  number={2},
  pages={227--236},
  year={2002},
  publisher={Elsevier}
}

@article{wierzba2023sources,
  title={Sources of variability in the syntactic flexibility of idioms},
  author={Wierzba, Marta and Brown, Jessica MM and Fanselow, Gisbert},
  journal={Glossa: a journal of general linguistics},
  volume={8},
  number={1},
  year={2023},
  publisher={Open Library of Humanities}
}

@article{sprenger2006lexical,
  title={Lexical access during the production of idiomatic phrases},
  author={Sprenger, Simone A and Levelt, Willem JM and Kempen, Gerard},
  journal={Journal of memory and language},
  volume={54},
  number={2},
  pages={161--184},
  year={2006},
  publisher={Elsevier}
}

@inproceedings{
olmo2,
title={2 {OLM}o 2 Furious ({COLM}{\textquoteright}s Version)},
author={Evan Pete Walsh and Luca Soldaini and Dirk Groeneveld and Kyle Lo and Shane Arora and Akshita Bhagia and Yuling Gu and Shengyi Huang and Matt Jordan and Nathan Lambert and Dustin Schwenk and Oyvind Tafjord and Taira Anderson and David Atkinson and Faeze Brahman and Christopher Clark and Pradeep Dasigi and Nouha Dziri and Allyson Ettinger and Michal Guerquin and David Heineman and Hamish Ivison and Pang Wei Koh and Jiacheng Liu and Saumya Malik and William Merrill and Lester James Validad Miranda and Jacob Morrison and Tyler Murray and Crystal Nam and Jake Poznanski and Valentina Pyatkin and Aman Rangapur and Michael Schmitz and Sam Skjonsberg and David Wadden and Christopher Wilhelm and Michael Wilson and Luke Zettlemoyer and Ali Farhadi and Noah A. Smith and Hannaneh Hajishirzi},
booktitle={Second Conference on Language Modeling},
year={2025},
url={https://openreview.net/forum?id=2ezugTT9kU}
}

@inproceedings{ententen,
  author    = {Jakubíček, Miloš and Kilgarriff, Adam and Kovář, Vojtěch and Rychlý, Pavel and Suchomel, Vít},
  title     = {The {TenTen} Corpus Family},
  booktitle = {Proceedings of the 7th International Corpus Linguistics Conference ({CL})},
  pages     = {125--127},
  year      = {2013},
  month     = jul
}

@inproceedings{stowe2022impli,
  title={IMPLI: Investigating NLI models’ performance on figurative language},
  author={Stowe, Kevin and Utama, Prasetya and Gurevych, Iryna},
  booktitle={Proceedings of the 60th Annual Meeting of the Association for Computational Linguistics (Volume 1: Long Papers)},
  pages={5375--5388},
  year={2022}
}

@inproceedings{devlin-etal-2019-bert,
    title = "{BERT}: Pre-training of Deep Bidirectional Transformers for Language Understanding",
    author = "Devlin, Jacob  and
      Chang, Ming-Wei  and
      Lee, Kenton  and
      Toutanova, Kristina",
    editor = "Burstein, Jill  and
      Doran, Christy  and
      Solorio, Thamar",
    booktitle = "Proceedings of the 2019 Conference of the North {A}merican Chapter of the Association for Computational Linguistics: Human Language Technologies, Volume 1 (Long and Short Papers)",
    month = jun,
    year = "2019",
    address = "Minneapolis, Minnesota",
    publisher = "Association for Computational Linguistics",
    url = "https://aclanthology.org/N19-1423/",
    doi = "10.18653/v1/N19-1423",
    pages = "4171--4186"
}

@inproceedings{modernBERT,
    title = "Smarter, Better, Faster, Longer: A Modern Bidirectional Encoder for Fast, Memory Efficient, and Long Context Finetuning and Inference",
    author = {Warner, Benjamin  and
      Chaffin, Antoine  and
      Clavi{\'e}, Benjamin  and
      Weller, Orion  and
      Hallstr{\"o}m, Oskar  and
      Taghadouini, Said  and
      Gallagher, Alexis  and
      Biswas, Raja  and
      Ladhak, Faisal  and
      Aarsen, Tom  and
      Adams, Griffin Thomas  and
      Howard, Jeremy  and
      Poli, Iacopo},
    editor = "Che, Wanxiang  and
      Nabende, Joyce  and
      Shutova, Ekaterina  and
      Pilehvar, Mohammad Taher",
    booktitle = "Proceedings of the 63rd Annual Meeting of the Association for Computational Linguistics (Volume 1: Long Papers)",
    month = jul,
    year = "2025",
    address = "Vienna, Austria",
    publisher = "Association for Computational Linguistics",
    url = "https://aclanthology.org/2025.acl-long.127/",
    doi = "10.18653/v1/2025.acl-long.127",
    pages = "2526--2547",
    ISBN = "979-8-89176-251-0"
}

@Article{Bulkes2017,
author={Bulkes, Nyssa Z.
and Tanner, Darren},
title={``Going to town'': Large-scale norming and statistical analysis of 870 American English idioms},
journal={Behavior Research Methods},
year={2017},
month={Apr},
day={01},
volume={49},
number={2},
pages={772-783},
issn={1554-3528},
doi={10.3758/s13428-016-0747-8},
url={https://doi.org/10.3758/s13428-016-0747-8}
}

@inproceedings{tenney-etal-2019-bert,
    title = "{BERT} Rediscovers the Classical {NLP} Pipeline",
    author = "Tenney, Ian  and
      Das, Dipanjan  and
      Pavlick, Ellie",
    editor = "Korhonen, Anna  and
      Traum, David  and
      M{\`a}rquez, Llu{\'i}s",
    booktitle = "Proceedings of the 57th Annual Meeting of the Association for Computational Linguistics",
    month = jul,
    year = "2019",
    address = "Florence, Italy",
    publisher = "Association for Computational Linguistics",
    url = "https://aclanthology.org/P19-1452/",
    doi = "10.18653/v1/P19-1452",
    pages = "4593--4601"
}

@inproceedings{jawahar-etal-2019-bert,
    title = "What Does {BERT} Learn about the Structure of Language?",
    author = "Jawahar, Ganesh  and
      Sagot, Beno{\^i}t  and
      Seddah, Djam{\'e}",
    editor = "Korhonen, Anna  and
      Traum, David  and
      M{\`a}rquez, Llu{\'i}s",
    booktitle = "Proceedings of the 57th Annual Meeting of the Association for Computational Linguistics",
    month = jul,
    year = "2019",
    address = "Florence, Italy",
    publisher = "Association for Computational Linguistics",
    url = "https://aclanthology.org/P19-1356/",
    doi = "10.18653/v1/P19-1356",
    pages = "3651--3657"
}

@article{Tabossi2009,
  author  = {Tabossi, Patrizia and Fanari, Rachele and Wolf, Kristine},
  title   = {Processing idiomatic expressions: Effects of semantic compositionality},
  journal = {Journal of Experimental Psychology: Learning, Memory, and Cognition},
  volume  = {35},
  number  = {2},
  pages   = {313--327},
  year    = {2009}
}

@book{Bybee_2010, place={Cambridge}, title={Language, Usage and Cognition}, publisher={Cambridge University Press}, author={Bybee, Joan}, year={2010}}

@book{10.1093/acprof:oso/9780199268511.001.0001,
    author = {Goldberg, Adele},
    title = {Constructions at Work: The Nature of Generalization in Language},
    publisher = {Oxford University Press},
    year = {2005},
    month = {12},
    isbn = {9780199268511},
    doi = {10.1093/acprof:oso/9780199268511.001.0001},
    url = {https://doi.org/10.1093/acprof:oso/9780199268511.001.0001},
}

@article{Riehemann:2001,
author = {Riehemann, Susanne},
year = {2001},
month = {01},
pages = {},
title = {A constructional approach to idioms and word formation}
}

@inproceedings{mi-etal-2025-rolling,
    title = "Rolling the {DICE} on Idiomaticity: How {LLM}s Fail to Grasp Context",
    author = "Mi, Maggie  and
      Villavicencio, Aline  and
      Moosavi, Nafise Sadat",
    editor = "Che, Wanxiang  and
      Nabende, Joyce  and
      Shutova, Ekaterina  and
      Pilehvar, Mohammad Taher",
    booktitle = "Proceedings of the 63rd Annual Meeting of the Association for Computational Linguistics (Volume 1: Long Papers)",
    month = jul,
    year = "2025",
    address = "Vienna, Austria",
    publisher = "Association for Computational Linguistics",
    url = "https://aclanthology.org/2025.acl-long.362/",
    doi = "10.18653/v1/2025.acl-long.362",
    pages = "7314--7332",
    ISBN = "979-8-89176-251-0"
}

@misc{olmo3,
      title={Olmo 3}, 
      author={{Team Olmo} and Allyson Ettinger and Amanda Bertsch and Bailey Kuehl and David Graham and David Heineman and Dirk Groeneveld and Faeze Brahman and Finbarr Timbers and Hamish Ivison and Jacob Morrison and Jake Poznanski and Kyle Lo and Luca Soldaini and Matt Jordan and Mayee Chen and Michael Noukhovitch and Nathan Lambert and Pete Walsh and Pradeep Dasigi and Robert Berry and Saumya Malik and Saurabh Shah and Scott Geng and Shane Arora and Shashank Gupta and Taira Anderson and Teng Xiao and Tyler Murray and Tyler Romero and Victoria Graf and Akari Asai and Akshita Bhagia and Alexander Wettig and Alisa Liu and Aman Rangapur and Chloe Anastasiades and Costa Huang and Dustin Schwenk and Harsh Trivedi and Ian Magnusson and Jaron Lochner and Jiacheng Liu and Lester James V. Miranda and Maarten Sap and Malia Morgan and Michael Schmitz and Michal Guerquin and Michael Wilson and Regan Huff and Ronan Le Bras and Rui Xin and Rulin Shao and Sam Skjonsberg and Shannon Zejiang Shen and Shuyue Stella Li and Tucker Wilde and Valentina Pyatkin and Will Merrill and Yapei Chang and Yuling Gu and Zhiyuan Zeng and Ashish Sabharwal and Luke Zettlemoyer and Pang Wei Koh and Ali Farhadi and Noah A. Smith and Hannaneh Hajishirzi},
      year={2025},
      eprint={2512.13961},
      archivePrefix={arXiv},
      primaryClass={cs.CL},
      url={https://arxiv.org/abs/2512.13961}, 
}

@article{InfiniGram,
  title={Infini-gram: Scaling Unbounded n-gram Language Models to a Trillion Tokens},
  author={Liu, Jiacheng and Min, Sewon and Zettlemoyer, Luke and Choi, Yejin and Hajishirzi, Hannaneh},
  journal={arXiv preprint arXiv:2401.17377},
  year={2024}
}

@ARTICLE{shannon-1948,
  author={Shannon, C. E.},
  journal={The Bell System Technical Journal}, 
  title={A mathematical theory of communication}, 
  year={1948},
  volume={27},
  number={3},
  pages={379-423},
  doi={10.1002/j.1538-7305.1948.tb01338.x}}

@Article{Gibbs1989,
author={Gibbs, Raymond W.
and Nayak, Nandini P.
and Cutting, Cooper},
title={How to kick the bucket and not decompose: Analyzability and idiom processing},
journal={Journal of Memory and Language},
year={1989},
month={Oct},
day={01},
volume={28},
number={5},
pages={576-593},
issn={0749-596X},
url={https://www.sciencedirect.com/science/article/pii/0749596X89900144}
}

@inproceedings{wasow1983idioms,
  title={Idioms: An interim report},
  author={Wasow, Thomas and Sag, Ivan and Nunberg, Geoffrey},
  booktitle={Proceedings of the XIIIth international congress of linguistics},
  volume={29},
  year={1983},
  organization={Tokyo}
}

@book{Nunberg1978Pragmatics,
  author    = {Nunberg, Geoffrey},
  title     = {The Pragmatics of Reference},
  year      = {1978},
  address   = {Bloomington, IN},
  publisher = {Indiana University Linguistics Club}
}

@article{gibbs_nayak_1989,
title = {Psycholinguistic studies on the syntactic behavior of idioms},
journal = {Cognitive Psychology},
volume = {21},
number = {1},
pages = {100-138},
year = {1989},
issn = {0010-0285},
doi = {https://doi.org/10.1016/0010-0285(89)90004-2},
url = {https://www.sciencedirect.com/science/article/pii/0010028589900042},
author = {Raymond W Gibbs and Nandini P Nayak}
}

@article{Chomsky_1980, title={Rules and representations}, volume={3}, DOI={10.1017/S0140525X00001515}, number={1}, journal={Behavioral and Brain Sciences}, author={Chomsky, Noam}, year={1980}, pages={1–15}}

@article{fraser1970idioms,
  title={Idioms within a transformational grammar},
  author={Fraser, Bruce},
  journal={Foundations of language},
  pages={22--42},
  year={1970},
  publisher={JSTOR}
}

@inbook {heringer,
      author = "James T. Heringer",
      title = "Idioms and Lexicalization in English",
      booktitle = "",
      year = "1976",
      publisher = "Brill",
      address = "Leiden, The Netherlands",
      isbn = "9789004368842",
      doi = "10.1163/9789004368842_008",
      pages=      "205 - 216",
      url = "https://brill.com/view/book/edcoll/9789004368842/BP000008.xml"
}

@incollection{Katz1973,
  author    = {Katz, Jerrold J.},
  title     = {Compositionality, Idiomaticity, and Lexical Substitution},
  booktitle = {A Festschrift for Morris Halle},
  editor    = {Anderson, Stephen R. and Kiparsky, Paul},
  pages     = {357--376},
  year      = {1973},
  address   = {New York},
  publisher = {Holt, Rinehart and Winston}
}

@article{Tabossi2009Idioms,
  author  = {Tabossi, Patrizia and Fanari, Rachele and Wolf, Karoline},
  title   = {Why are idioms recognized fast?},
  journal = {Memory \& Cognition},
  year    = {2009},
  volume  = {37},
  number  = {4},
  pages   = {529--540},
  doi     = {10.3758/MC.37.4.529}
}

@article{gibbs1989speakers,
  title={Speakers' assumptions about the lexical flexibility of idioms},
  author={Gibbs, Raymond W and Nayak, Nandini P and Bolton, John L and Keppel, Melissa E},
  journal={Memory \& cognition},
  volume={17},
  number={1},
  pages={58--68},
  year={1989},
  publisher={Springer}
}

@article{Titone_norms,
author = {Debra A. Titone and Cynthia M. Connine},
title = {Descriptive Norms for 171 Idiomatic Expressions: Familiarity, Compositionality, Predictability, and Literality},
journal = {Metaphor and Symbolic Activity},
volume = {9},
number = {4},
pages = {247--270},
year = {1994},
publisher = {Routledge},
doi = {10.1207/s15327868ms0904\_1},
URL = { 
        https://doi.org/10.1207/s15327868ms0904_1
},
eprint = { 
        https://doi.org/10.1207/s15327868ms0904_1
}
}

@article{Titone1999,
title = {On the compositional and noncompositional nature of idiomatic expressions},
journal = {Journal of Pragmatics},
volume = {31},
number = {12},
pages = {1655-1674},
year = {1999},
note = {Literal and Figurative Language},
issn = {0378-2166},
doi = {https://doi.org/10.1016/S0378-2166(99)00008-9},
url = {https://www.sciencedirect.com/science/article/pii/S0378216699000089},
author = {Debra A. Titone and Cynthia M. Connine}
}

@inproceedings{kornblith2019similarity,
  title={Similarity of neural network representations revisited},
  author={Kornblith, Simon and Norouzi, Mohammad and Lee, Honglak and Hinton, Geoffrey},
  booktitle={International conference on machine learning},
  pages={3519--3529},
  year={2019},
  organization={PMlR}
}

@article{Kantorovitch1958,
  title={On the Translocation of Masses},
  author={L. Kantorovitch},
  journal={Management Science},
  year={1958},
  volume={5},
  pages={1-4},
  url={https://api.semanticscholar.org/CorpusID:214798034}
}

@article{Vaserstein1969Markov,
  author  = {Vaserstein, L. N.},
  title   = {Markov Processes over Denumerable Products of Spaces, Describing Large Systems of Automata},
  journal = {Problemy Peredachi Informatsii},
  year    = {1969},
  volume  = {5},
  number  = {3},
  pages   = {64--72},
  url     = {http://mi.mathnet.ru/ppi1811},
  mrnumber = {314115},
  zbl     = {0273.60054},
  language = {russian}
}

@article{CacciariLevorato1998,
  author    = {Cacciari, Cristina and Levorato, Maria Chiara},
  title     = {The effect of semantic analyzability of idioms in metalinguistic tasks},
  journal   = {Metaphor and Symbol},
  volume    = {13},
  number    = {3},
  pages     = {159--177},
  year      = {1998},
  publisher = {Lawrence Erlbaum},
  doi       = {10.1207/s15327868ms1303_1},
  issn      = {1532-7868}
}

@book{gibbs2012interpreting,
  title={Interpreting figurative meaning},
  author={Gibbs Jr, Raymond W and Colston, Herbert L},
  year={2012},
  publisher={Cambridge University Press}
}

@inproceedings{vaswani_attention,
 author = {Vaswani, Ashish and Shazeer, Noam and Parmar, Niki and Uszkoreit, Jakob and Jones, Llion and Gomez, Aidan N and Kaiser, \L ukasz and Polosukhin, Illia},
 booktitle = {Advances in Neural Information Processing Systems},
 editor = {I. Guyon and U. Von Luxburg and S. Bengio and H. Wallach and R. Fergus and S. Vishwanathan and R. Garnett},
 pages = {},
 publisher = {Curran Associates, Inc.},
 title = {Attention is All you Need},
 url = {https://proceedings.neurips.cc/paper_files/paper/2017/file/3f5ee243547dee91fbd053c1c4a845aa-Paper.pdf},
 volume = {30},
 year = {2017}
}

@article{Vulchanova2019Boon,
  title     = {Boon or burden? The role of compositional meaning in figurative language processing and acquisition},
  author    = {Vulchanova, Mila and Milburn, Evelyn and Vulchanov, Valentin and Baggio, Giosu{\`e}},
  journal   = {Journal of Logic, Language and Information},
  year      = {2019},
  volume    = {28},
  number    = {2},
  pages     = {359--387},
  publisher = {Springer},
  address   = {Germany},
  doi       = {10.1007/s10849-019-09282-7},
  issn      = {1572-9583}
}

@article{CACCIARI1988,
title = {The comprehension of idioms},
journal = {Journal of Memory and Language},
volume = {27},
number = {6},
pages = {668-683},
year = {1988},
issn = {0749-596X},
doi = {https://doi.org/10.1016/0749-596X(88)90014-9},
url = {https://www.sciencedirect.com/science/article/pii/0749596X88900149},
author = {Cristina Cacciari and Patrizia Tabossi}
}

@inproceedings{mi-etal-2025-input,
    title = "From Input Perception to Predictive Insight: Modeling Model Blind Spots Before They Become Errors",
    author = "Mi, Maggie  and
      Villavicencio, Aline  and
      Moosavi, Nafise Sadat",
    editor = "Christodoulopoulos, Christos  and
      Chakraborty, Tanmoy  and
      Rose, Carolyn  and
      Peng, Violet",
    booktitle = "Proceedings of the 2025 Conference on Empirical Methods in Natural Language Processing",
    month = nov,
    year = "2025",
    address = "Suzhou, China",
    publisher = "Association for Computational Linguistics",
    url = "https://aclanthology.org/2025.emnlp-main.1740/",
    doi = "10.18653/v1/2025.emnlp-main.1740",
    pages = "34328--34341",
    ISBN = "979-8-89176-332-6"
}

@inproceedings{sag_pain,
    author = {Sag, Ivan A. and Baldwin, Timothy and Bond, Francis and Copestake, Ann A. and Flickinger, Dan},
    title = {Multiword Expressions: A Pain in the Neck for NLP},
    year = {2002},
    isbn = {3540432191},
    publisher = {Springer-Verlag},
    address = {Berlin, Heidelberg},
    booktitle = {Proceedings of the Third International Conference on Computational Linguistics and Intelligent Text Processing},
    pages = {1–15},
    numpages = {15},
    series = {CICLing '02}
    }

@inproceedings{hale-2001-probabilistic,
    title = "A Probabilistic {E}arley Parser as a Psycholinguistic Model",
    author = "Hale, John",
    booktitle = "Second Meeting of the North {A}merican Chapter of the Association for Computational Linguistics",
    year = "2001",
    url = "https://aclanthology.org/N01-1021/"
}

@article{Levy2008Expectation,
  title        = {Expectation-based syntactic comprehension},
  author       = {Levy, Roger},
  journal      = {Cognition},
  volume       = {106},
  number       = {3},
  pages        = {1126--1177},
  year         = {2008},
  month        = mar,
  issn         = {0010-0277},
  doi          = {10.1016/j.cognition.2007.05.006},
  url          = {https://www.sciencedirect.com/science/article/pii/S0010027707001436}
}

@article{strassler1982idioms,
  title={Idioms in English},
  author={Strassler, Jurg},
  journal={A Pragmatic Analysis. Tubingen: Narr},
  year={1982}
}

@article{nunberg1994idioms,
  title={Idioms},
  author={Nunberg, Geoffrey and Sag, Ivan A and Wasow, Thomas},
  journal={Language},
  volume={70},
  number={3},
  pages={491--538},
  year={1994},
  publisher={Linguistic Society of America}
}

@article{libben2008multidetermined,
  title={The multidetermined nature of idiom processing},
  author={Libben, Maya R and Titone, Debra A},
  journal={Memory \& cognition},
  volume={36},
  number={6},
  pages={1103--1121},
  year={2008},
  publisher={Springer}
}

@article{chang_word_learning,
    author = {Chang, Tyler A. and Bergen, Benjamin K.},
    title = {Word Acquisition in Neural Language Models},
    journal = {Transactions of the Association for Computational Linguistics},
    volume = {10},
    pages = {1-16},
    year = {2022},
    month = {01},
    issn = {2307-387X},
    doi = {10.1162/tacl_a_00444},
    url = {https://doi.org/10.1162/tacl_a_00444},
    eprint = {https://direct.mit.edu/tacl/article-pdf/doi/10.1162/tacl_a_00444/1986589/tacl_a_00444.pdf},
}

@misc{openai2022chatgpt,
  author       = {{OpenAI}},
  title        = {ChatGPT: Optimizing Language Models for Dialogue},
  year         = {2022},
  howpublished = {\url{https://openai.com/blog/chatgpt}},
  note         = {Accessed: 2026-01-05}
}

@misc{llama,
      title={LLaMA: Open and Efficient Foundation Language Models}, 
      author={Hugo Touvron and Thibaut Lavril and Gautier Izacard and Xavier Martinet and Marie-Anne Lachaux and Timothée Lacroix and Baptiste Rozière and Naman Goyal and Eric Hambro and Faisal Azhar and Aurelien Rodriguez and Armand Joulin and Edouard Grave and Guillaume Lample},
      year={2023},
      eprint={2302.13971},
      archivePrefix={arXiv},
      primaryClass={cs.CL},
      url={https://arxiv.org/abs/2302.13971}, 
}

@article{hubbard2023predictability,
  title={Predictability and decomposability separately contribute to compositional processing of idiomatic language},
  author={Hubbard, Ryan and Bulkes, Nyssa and Lai, Vicky Tzuyin},
  journal={Psychophysiology},
  volume={60},
  number={8},
  pages={e14269},
  year={2023},
  publisher={Wiley Online Library}
}

@article{nordmann2014,
title = {Familiarity breeds dissent: Reliability analyses for British-English idioms on measures of familiarity, meaning, literality, and decomposability},
journal = {Acta Psychologica},
volume = {149},
pages = {87-95},
year = {2014},
note = {Including Special section articles of Temporal Processing Within and Across Senses - Part-2},
issn = {0001-6918},
doi = {https://doi.org/10.1016/j.actpsy.2014.03.009},
url = {https://www.sciencedirect.com/science/article/pii/S0001691814000845},
author = {Emily Nordmann and Alexandra A. Cleland and Rebecca Bull}
}

@inproceedings{ficarra-etal-2025-distributional,
    title = "A Distributional Perspective on Word Learning in Neural Language Models",
    author = "Ficarra, Filippo  and
      Cotterell, Ryan  and
      Warstadt, Alex",
    editor = "Chiruzzo, Luis  and
      Ritter, Alan  and
      Wang, Lu",
    booktitle = "Proceedings of the 2025 Conference of the Nations of the Americas Chapter of the Association for Computational Linguistics: Human Language Technologies (Volume 1: Long Papers)",
    month = apr,
    year = "2025",
    address = "Albuquerque, New Mexico",
    publisher = "Association for Computational Linguistics",
    url = "https://aclanthology.org/2025.naacl-long.558/",
    doi = "10.18653/v1/2025.naacl-long.558",
    pages = "11184--11207",
    ISBN = "979-8-89176-189-6"
}

@article{sheinfux2019verbal,
  title={Verbal multiword expressions: Idiomaticity and flexibility},
  author={Sheinfux, Livnat Herzig and Greshler, Tali and Melnik, Nurit and Winter, Shuly},
  journal={Representation and parsing of multiword expressions: Current trends},
  volume={3},
  pages={35},
  year={2019},
  publisher={Language Science Press}
}

@incollection{cacciari_1991,
title = {Chapter 9 Understanding Idiomatic Expressions: The Contribution of Word Meanings},
editor = {Greg B. Simpson},
series = {Advances in Psychology},
publisher = {North-Holland},
volume = {77},
pages = {217-240},
year = {1991},
booktitle = {Understanding Word and Sentence},
issn = {0166-4115},
doi = {https://doi.org/10.1016/S0166-4115(08)61535-6},
url = {https://www.sciencedirect.com/science/article/pii/S0166411508615356},
author = {Cristina Cacciari and Sam Glucksberg}
}

@Inbook{partee_1995,
author={Partee, Barbara H.},
title={Lexical semantics and compositionality.},
series={An invitation to cognitive science.},
year={1995},
publisher={The MIT Press},
address={Cambridge,  MA,  US},
pages={311-360},
isbn={0-262-15044-1 (Hardcover); 0-262-65044-4 (Paperback)}
}

@article{dolev2025decomposability,
  title={Decomposability and the syntactic flexibility of Hebrew idioms},
  author={Dolev, Tom and Siloni, Tal},
  journal={Glossa: a journal of general linguistics},
  volume={10},
  number={1},
  year={2025},
  publisher={Open Library of Humanities}
}
